\documentclass[acmsmall, nonacm, screen]{acmart}

\usepackage{amsmath}
\usepackage{bbm}
\usepackage{multirow}
\usepackage{subfigure}
\usepackage{hyperref}
\usepackage{breqn}
\usepackage{mathtools}
\usepackage{microtype}
\usepackage{cleveref}
\usepackage{lscape}
\usepackage{upgreek}
\usepackage{graphicx}
\usepackage{booktabs}
\usepackage{caption}
\usepackage{amsthm}
\usepackage{algorithm}
\usepackage{algpseudocode}
\usepackage{xcolor}
\usepackage[detect-all]{siunitx}
\usepackage[compact]{titlesec}

\definecolor{_yellow}{rgb}{0.94, 0.86, 0.51}
\definecolor{_red}{rgb}{0.89, 0.44, 0.48}
\definecolor{_blue}{rgb}{0.63,0.79,0.95}
\definecolor{_green}{rgb}{0.64,0.76,0.68}
\definecolor{mark}{rgb}{1.0, 0.03, 0.0}

\newcommand{\method}{\texttt{FedLN}}
\renewcommand{\shortauthors}{Tsouvalas et al.}

\DeclareMathOperator*{\argmax}{arg\,max}

\DeclareMathOperator{\diag}{diag}
\algnewcommand{\IIf}[1]{\State\algorithmicif\ #1\ \algorithmicthen}
\algnewcommand{\EndIIf}{\unskip\ \algorithmicend\ \algorithmicif}

\theoremstyle{definition}
\newtheorem{definition}{Definition}[section]

\begin{document}
\title[Federated Learning with Noisy Labels]{Labeling Chaos to Learning Harmony: Federated Learning with Noisy Labels}

\author{Vasileios Tsouvalas}
\affiliation{%
\institution{Eindhoven University of Technology}
\city{Eindhoven}
\country{The Netherlands}}
\email{v.tsouvalas@tue.nl}

\author{Aaqib Saeed}
\affiliation{%
\institution{Eindhoven University of Technology}
\city{Eindhoven}
\country{The Netherlands}}
\email{a.saeed@tue.nl}

\author{Tanir Ozcelebi}
\affiliation{%
\institution{Eindhoven University of Technology}
\city{Eindhoven}
\country{The Netherlands}}
\email{t.ozcelebi@tue.nl}

\author{Nirvana Meratnia}
\affiliation{%
\institution{Eindhoven University of Technology}
\city{Eindhoven}
\country{The Netherlands}}
\email{n.meratnia@tue.nl}

\begin{abstract}
Federated Learning (FL) is a distributed machine learning paradigm that enables learning models from decentralized private datasets, where the labeling effort is entrusted to the clients. While most existing FL approaches assume high-quality labels are readily available on users' devices; in reality, label noise can naturally occur in FL and is closely related to clients' characteristics. Due to scarcity of available data and significant label noise variations among clients in FL, existing state-of-the-art centralized approaches exhibit unsatisfactory performance, while prior FL studies rely on excessive on-device computational schemes or additional clean data available on server. Here, we propose~\method, a framework to deal with label noise across different FL training stages; namely, FL initialization, on-device model training, and server model aggregation, able to accommodate the diverse computational capabilities of devices in a FL system. Specifically,~\method~computes per-client noise-level estimation in a single federated round and improves the models' performance by either correcting or mitigating the effect of noisy samples. Our evaluation on various publicly available vision and audio datasets demonstrate a $22$\% improvement on average compared to other existing methods for a label noise level of $60$\%. We further validate the efficiency of~\method~in human-annotated real-world noisy datasets and report a $4.8$\% increase on average in models' recognition performance, highlighting that~\method~can be useful for improving FL services provided to everyday users.
\end{abstract}

\keywords{federated learning, noisy labels, label correction, deep learning, knowledge distillation}

\maketitle

\section{Introduction}\label{introduction}

Recent advances in smartphones, wearables, and the Internet of Things devices have led to the continuous generation of massive amounts of data from embedded sensors and users' interactions with various applications. The ubiquity of these devices and the exponential growth of the data they produce present a significant opportunity to tackle critical problems in domains such as healthcare, well-being, and manufacturing. Traditionally, machine learning (ML) approaches require the distributed data to be stored or aggregated in a centralized cloud-based server before being further processed to solve a specific problem. However, the rapidly increasing volume of generated data, combined with high communication costs and bandwidth limitations, makes centralized data aggregation infeasible~\citep{9084352}. Furthermore, such centralized schemes may also be restricted by privacy issues and regulations, such as the General Data Protection Regulation (GDPR)\footnote{\href{https://gdpr-info.eu/}{https://gdpr-info.eu/}}.

To this end, the field of Federated Learning (FL)~\citep{fl} aims to enable distributed training of machine learning models on decentralized data residing on personal devices like smartphones and wearables. The key idea behind FL is to bring the computation closer to where the data resides to extensively harness data locality. In a FL regime, updates to the deep neural network models (e.g., their learnable parameters) are performed entirely on-device and communicated to the central server, which aggregates these updates from all participating devices to produce a unified global model. Unlike the standard centralized way of learning models, the salient differentiating factor of FL is that the data never leaves the user’s device, making it an appealing property for privacy-sensitive data. Recently, FL has been successfully applied to a wide range of tasks with great success~\citep{KeywordSpottingIID, KeywordSuggestion, fedstar}. Nevertheless, a common limitation of existing supervised FL approaches is the implicit assumption that on-device data are perfectly annotated~\citep{cifarn}. 

In reality, the quality of labeled data can vary depending on the data collection and annotation process. Under centralized regimes, having access to a larger and diverse pool of labeled data allows for better label validation and higher quality annotations than in federated settings~\citep{4,5,8}. Moreover, access to centrally aggregated data enables various label correction processes to be exploited to further improve the quality of labeled data, such as crowdsourcing, outsourcing and expert annotation. In contrast, the decentralized nature of FL hinders the collection of high-quality labeled data, leaving no way to verify the quality of labels. Here, the data annotation is typically performed through user interaction or automatically via a programmatic labeling functions, such as those used for keyboard query suggestions~\citep{KeywordSuggestion}. However, such techniques often result in noisy labels being assigned to the data samples either due to missing expertize of users, or due to the inherently noisy labels constructed from automatic labeling systems, such as ``\textit{weak}'' labeling~\citep{weak_labels}. Therefore, in FL, the presence of mislabeled data samples, referred to as label noise or noisy labels, can naturally occur, while there is no straightforward way to perform label correction.

The problem of training models under label noise has received noticeable attention with various proposed algorithms over the years~\citep{ls,cl,bTL}, and has emerged as a major practical challenge in the context of federated learning in recent times~\citep{2,4,5,6,7,8,9}. The non-i.i.d. nature of data in FL, which is characterized by variations in the data distribution across clients, can affect both the presence and distribution of label noise. Specifically, label noise in FL is closely related to the characteristics of the clients' devices and the expertise of their users. These unique label noise characteristics in FL make it challenging to successfully apply centralized learning schemes that filter noisy samples or mitigate their effect through regularization techniques~\citep{4,5,8}. To deal with label noise in FL, recent approaches often rely on repeated server-side aid, either in the form of additional clean data~\citep{8,9,10} or communication of client-sensitive data~\citep{2,5}, while computational expensive approaches have been proposed to perform label correct in FL~\citep{4}.

To the best of our knowledge, our work represents the first attempt to reduce the effect of label noise in the federated setting for various classification tasks in multiple stages of FL (i.e., initialization, local training step, server-side model aggregation) without relying on any additional server-side clean data, and with varying degree of compute requirements; thus, providing suitable solutions across a wide range of devices (low to high-end computation devices). We present three distinct FL schemes, namely NNC, AKD and NA-FedAvg, each tackling the problem of label noise in the federated setting in two stages: firstly, computing a per-client noise level estimation, and secondly, exploiting this knowledge to efficiently train deep neural networks to improve the performance of a given task, while correcting (or limiting the effect of) noisy labeled samples. Concisely, the main contributions of our work are as follows:

\begin{itemize}
    \item We propose a framework, called~\method\footnote{\href{https://github.com/FederatedML/FedLN}{https://github.com/FederatedML/FedLN}}~(Federated Learning with Label Noise) to address the generalizability issues introduced when training federated models using noisy labeled data.

    \item We design simple yet effective approaches to accurately estimate a per-client label noise level and identify clients with clean or relatively high-quality labels.
    
    \item We devise various mechanisms to alleviate or correct noisy labeled instances on a per-client basis, thus mitigating the need of user interaction for high-quality label acquisition.

    \item We demonstrate that our framework is highly useful for learning generalizable models under a variety of federated and label noise settings on diverse public datasets from both vision and audio domains, namely CIFAR-10~\citep{cifar10}, FashionMNIST~\citep{fmnist}, PathMNIST~\citep{pathmnist}, EuroSAT~\citep{eurosat}, and SpeechCommands~\citep{spcm}.

    \item We show that~\method~can improve recognition rate by $22$\% on average across all datasets compared to the fully-supervised federated model, when 60\% of labeled data contain noisy labels. Further evaluation of~\method~on real-world human annotated noisy datasets, namely CIFAR-10N/100N~\citep{cifarn}, exhibits an increase in recognition rate by $9$\% compared to the naive FL strategy.

\end{itemize}

\section{Background}\label{background}
In this section, we provide a brief overview of the statistical properties of label noise, the procedure of training a neural network with label noise, and FL to provide a foundation for our approach for training deep learning models with noisy labels in the federated setting.

\subsection{Label Noise}\label{ssec:label_noise}

Label noise refers to the misalignment between a ground truth label $y^{*}$ and an observed label $y$ in a given dataset. Specifically, in a $\mathcal{C}$-way classification problem, where $\mathcal{C}$ is the number of label categories, label noise can be considered as a class-conditional label flipping process $f\left(\cdot\right)$, which projects $y^{*} \rightarrow y$, in a way that every label in class $j\in\mathcal{C}$ may be independently mislabeled as class $i\in\mathcal{C}$ with probability $p\left(y{=}i{\mid}y^{*}{=}j\right)$, written in a shorthand notation as $p\left(y{\mid}y^{*}\right)$. Hence, in our work we assume that the occurrences of label noise are data-independent, i.e., $p\left(y{\mid}y^{*}{,}x\right) = p\left(y{\mid}y^{*}\right)$, similar to~\citep{bg1}. From the definition of label noise function $f\left(y^{*}{,}\mathcal{C}\right)$, a $\mathcal{C}\times\mathcal{C}$ noise distribution matrix denoted by $\mathcal{Q}_{y{\mid}y^{*}}$ can be defined, where each column corresponds to the probability distribution for an input instance with ground truth label $y^{*}{=}i$ to be assigned to label $j$.

Given the definitions above, we can characterize label noise through two statistic parameters, i.e., noise level (denoted by $n_{l}$), and noise sparsity (denoted by $n_{s}$). Inspired by~\citep{cl}, we provide a formal definition for each of these parameters, as follows:

\begin{definition}[\textbf{Noise Level}\label{def:noise_lvl}]
Noise level ($n_l$) quantifies the amount of label noise present in a given dataset. It is defined as the reverse probability of the sum along the diagonal of $\mathcal{Q}_{y{\mid}y^{*}}$, denoted as $ n_{l} = 1 - \diag\left(\mathcal{Q}_{y{\mid}y^{*}}\right)$. Intuitively, noise level of zero corresponds to a ``\textit{clean}'' dataset, where all observed labels match their ground truth labels, while a noise level of one can be considered as completely ``\textit{noisy}'' dataset.
\end{definition}

\begin{definition}[\textbf{Noise Sparsity}\label{def:noise_spar}]
Noise sparsity ($n_s$) quantifies the shape of the label noise present in a given dataset. It is defined by the probability concentration of noise in each column of $\mathcal{Q}_{y{\mid}y^{*}}$ when off-diagonals values are discarded. Thus, a high noise sparsity value indicates a non-uniformity of label noise, common in most real-world datasets. For example, a high-sparsity noise can indicate a confusion between classes that are perceived to be related by humans, i.e., mislabeling of a cat as tiger or a lion, rather than a cat as a bird or dog. Alternatively, zero level of noise sparsity corresponds to completely random noise, where all instances belonging to one class can be confused with any other class. The special case of ``\textit{class-flipping}'' can be constructed for $n_{s}=1$, where instances that belong to a pair of two classes are confused. In this case, we consider the noise probabilities between these pair of classes to be equal, i.e., $p\left(y{=}i{\mid}y^{*}{=}j\right) = p\left(y{=}j{\mid}y^{*}{=}i\right)$. While in reality non-diagonal entries in $\mathcal{Q}_{y{\mid}y^{*}}$ are non-zero (label noise is essentially unavoidable among classes), we allow zero values to be present in $\mathcal{Q}_{y{\mid}y^{*}}$ and consider $n_s$ as the fraction of positive non-diagonal entries per column of $\mathcal{Q}_{y{\mid}y^{*}}$, similar to~\citep{cl}.

\end{definition}

\subsection{Federated Learning}\label{ssec:fl}

Federated Learning is a collaborative learning paradigm that aims to learn a single, global model from data stored on remote clients with no need to share their data with a central server. Specifically, with the data residing on clients' devices, a subset of clients is selected to perform a number of local SGD steps on their data in parallel on each communication round. Upon completion, clients exchange their models' weights updates with the server, aiming to learn a unified global model by aggregating these updates. Formally, the goal of FL is typically to minimize the following objective function:

\begin{equation} \label{eqn:FL}
     \min_{\theta} \mathcal{L}_{\theta} = \sum_{m=1}^{M} \gamma_{m} {\mathcal{L}}_m(\theta),
\end{equation}

\noindent where $\mathcal{L}_m$ is the loss function of the $m^{th}$ client and $\gamma_{m}$ corresponds to the relative impact of the $m^{th}$ client on the construction of the global model. For the FedAvg~\citep{fedavg} algorithm, $\gamma_{m}$ is equal to the ratio of client's local data $N_m$ over all training samples, i.e., $\left (\gamma_{m} = \frac{N_m}{N}\right)$.

\subsubsection*{\textbf{Federated Label Noise}} 
Considering the traditional centralized learning, noise distribution can be characterized by a single noise distribution matrix $\mathcal{Q}_{y{\mid}y^{*}}$. However, in FL, where data is fragmented across multiple clients, distinct noise distributions among clients have to be considered, as label noise is closely related to the clients' characteristics, i.e., user's expertise or preferences. Subsequently, in FL, noise distribution matrices among clients can differ significantly, i.e., $\mathcal{Q}_{y{\mid}y^{*}}^{i} \neq \mathcal{Q}_{y{\mid}y^{*}}^{j}$ with $i{,}j$ indicating any pair of clients. These naturally occurring differences in noise distributions among clients (i.e., clients' noise profiles) can introduce additional challenges for the FL process, especially during models' aggregation step. 

\subsection{Learning from Noisy Labels}\label{ml_label_noise}

The goal of a $\mathcal{C}$-way supervised learning task is to learn a function that maps an input instance $x$ to a corresponding ground truth label $y_{i}^{*} ~\epsilon \left \{ 1, \cdots, \mathcal{C} \right \}$. Let $p_\theta\left(y \mid x \right)$ be a neural network that is parameterized by weights $\theta$ that predicts softmax outputs $y$ for a given input $x$. In a typical classification problem, the model is provided with a training dataset $\mathcal{D} = \left \{ \left ( x_{i},y_{i}^{*} \right ) \right \}_{i=1}^{N}$, and aims to minimize the following objective function by learning the model's parameters $\theta$:

\begin{equation} \label{eqn:clean_learn}
    \mathcal{L}_{\theta}\left(\mathcal{D}\right) = \mathit{l}\left(y^{*}{,}p_\theta\left(y{\mid}x\right)\right),
\end{equation}

\noindent where $\mathcal{L}_{\theta}\left(\mathcal{D}\right)$ is the minimization function for supervised learning on $\mathcal{D}$, $\mathit{l}\left(\cdot\right)$ denotes a certain loss, and $p_\theta$ is the neural network that is parameterized by weights $\theta$.

In real-world scenarios, in which data labels are often noisy, the neural network $p_\theta\left(y{\mid}x\right)$ is trained on noisy labels $y$, instead of the actual ground-truth labels $y^{*}$. Similar to Equation~\ref{eqn:clean_learn}, the objective is to minimize $\mathcal{L}_{\theta}$ on $\mathcal{D}_{n} = \{ \left(x_{i}{,}y_{i}\right ) \}_{i=1}^{N_n}$ by learning the model's parameters $\theta$, as follows:

\begin{equation} \label{eqn:noisy_learn}
    \mathcal{L}_{\theta}\left(\mathcal{D}_{n}\right) = \mathit{l}\left(y{,}p_\theta\left(y{\mid}x\right)\right),
\end{equation}

Here, noisy labeled instances interference with the loss minimization process, since the computed loss is over the noisy dataset $\mathcal{D}_{n}$. Specifically, the neural network $p_{\theta}$ can easily memorize noisy labels and consequently degenerate network's generalization on unseen data~\citep{noises1}.

\section{Related Work}\label{related_work}

\subsubsection*{Noisy Label Learning} Supervised deep learning approaches primarily use data-label pairs to train models, prompting extensive research on neural network robustness against noisy labels. Here, researchers have focused on tackling label noise through loss correction techniques, which involve adjusting the loss value per sample to mitigate the impact of noisy samples. Early techniques performed loss correction by estimating the noise distribution matrix, either via pre-trained models~\citep{patrini} or by utilizing a clean validation set~\citep{bengio}. In addition to estimating the noise distribution matrix,~\citep{scce} proposed the use of symmetric cross-entropy to help deal with noisy labels. Furthermore,~\citep{bTL} utilized a generalized cross-entropy loss with heavy-tailed softmax probabilities via two tunable parameters to limit the loss value per sample; thus, minimizing the effect of noisy labels. Regularization techniques, such as label smoothing~\citep{ls, ls_rate}, and data augmentation techniques, like MixUp~\citep{mixup}, show promise in handling label noise. Co-learning~\citep{tan2021co} involves performing supervised and self-supervised learning in a cooperative way to prevent noisy label memorization. Alternatively, direct estimation and removal of noisy labeled instances prior to training have been explored~\citep{gmm_noise,cl}.~\citep{gmm_noise} utilizes Gaussian Mixture Models (GMM) based on gradient values, while Confidence Learning~\citep{cl} is based on probabilistic thresholds from pre-trained neural network predictions. Based on the ``\textit{memorization effects}'' of deep networks, where models fit data with clean labels prior to noisy instances~\citep{mem}, researchers have proposed early-stopping mechanisms to mitigate the negative effects of label noise in models generalization~\citep{es1,es2,es3,patrini}. Nevertheless, in the federated setting, noise patterns vary across clients and on-device data can be scarce, introducing further complexities. This necessitates modeling noise profiles, assessing the model's confidence in its loss values, and determining appropriate thresholds to detect noisy instances~\citep{4,5,9}. Our work performs on a thorough evaluation of centralized approaches that deal with label noise in the federated setting. By examining these techniques, we aim to enable future research advancements in the area of federated learning under the presence of label noise.

\subsubsection*{Label Noise in Federated Learning} Recent research has addressed the issue of label noise in the federated learning setting~\citep{2,4,5,6,7,8,9}. Approaches such as~\citep{2,5} focus on filtering noisy samples. For example,\citep{2} utilizes communication of class-wise data centroids among clients to construct decision boundaries across classes, while~\citep{5} relies on the communication of data features to the server for identifying noise instances. In addition to data filtering, label correction techniques have been investigated in federated learning~\citep{4,6,7}. CLC~\citep{7} applies consensus-based label correction technology, enabling clients to cooperate in correcting labels through a consensus mechanism. Fedcorr~\citep{4} employs a multi-stage scheme, where clean samples are first detected using GMM based on loss scores, and then used to train a model that provides pseudo-labels for the noisy instances. Furthermore,~\citep{6} proposes the use of meta-learning to jointly learn the underlying recognition task and the noise distribution matrix, mapping noisy labeled instances to their correct counterparts during training. Recently,~\citep{8,9,10} explored the utilization of a small clean dataset to quantify the credibility of on-device data and adjust the weighted aggregation process accordingly. However, a common limitation of these approaches is the significant computational overhead they often require for performing label correction. Alternatively, approaches such as~\citep{2,5} necessitate the communication of user-related information and assume that clients share the same noise ratio for effective data filtering. Moreover, for efficient label noise handling with a low computational footprint on clients, approaches such as~\citep{5,8,9,10} often assume the availability of additional clean data. To address the aforementioned problems, we propose~\method, a framework that provides simple yet effective approaches to accurately estimate label noise on a per-client basis and offer robust learning schemes to learn better generalizable federated models under the presence of label noise without relying on additional clean data, complex learning schemes, or communication of client-sensitive data.

\section{Methodology}\label{methodology}

In this section, we present our federated learning framework,~\method, for FL models under the presence of label noise. Firstly, we provide a formal definition of the underlying problem. Then, we discuss our proposed techniques for determining a per-client noise level estimation in detail. Finally, we provide a thorough description of our developed approached for handling and mitigating the impact of noisy labels during training of deep models in the federated setting.

\subsection{Problem Formulation}

We focus on the problem of federated learning with noisy labels, where clients' data samples are often mislabeled either due to missing expertise of annotators, users' mistakes, or error in the automated procedure for label inference. While noise can be present in input-space of data, in this work, we solely focus on ``\textit{noise}" to be present in the label-space (i.e., categorize in case of classification problems). In particular, as label noise, we consider the misalignment between a ground truth label $y^{*}$ and an observed label $y$ in a given dataset, which is characterized by noise level $n_{l}$ (reverse probability of the sum along the diagonal of $\mathcal{Q}_{y{\mid}y^{*}}$) and $n_{s}$ (fraction of non-zeros entries per column of $\mathcal{Q}_{y{\mid}y^{*}}$). With label noise can be closely related to the clients' characteristics in FL, distinct noise distribution matrices (i.e., clients' noise profiles) can exist among clients. These varying label noise profiles introduce risks of overfitting to noisy data, negatively impacting the server-side model aggregation, while necessitate the need to detect label noise on a client-based level. Additionally, clients holding data with high-quality labels, i.e., $n_{l} \leqslant \varepsilon$ with $\varepsilon \rightarrow 0$, may be present in the FL process. With~\method, we aim to eliminate the effect of on-device mislabeled examples in the training process and to improve the performance of FL models, alleviating the common assumption that clients hold well-annotated data. 

Formally in FL, we have a set of $M$ clients, each holding a training set $\mathcal{D}^{m}$. Subsequently, each client's dataset, $\mathcal{D}^{m}$, can be divided into a correctly labeled set (clean data) $\mathcal{D}_{c}^{m} = \left \{ \left ( x_{i},y_{i}^{*} \right ) \right \}_{i=1}^{N_{c}^{m}}$ and a noisy labeled set (noisy data) $\mathcal{D}_{n}^{m} = \{ \left ( x_{i}, y_{i} \right )\}_{i=1}^{N_{n}^{m}}$, where $N^{m} = N_{c}^{m} + N_{n}^{m}$ is the total number of data samples stored on the $\mathit{m^{th}}$ client. The label noise level present in the $\mathit{m^{th}}$ client's data is defined by $n_{l}^{m} = \frac{N_{n}^{m}}{N^{m}}$ and $N = \sum_{i=0}^{M} N^{m}$ is the total number of samples present during training. We aim to learn a global unified model $G$ without clients sharing any of their local data ($\mathcal{D}^m$), while minimizing the effect of noisy label set $\mathcal{D}_{n}^{m}$ on the training process. Specifically, the objective function we aim to minimize is the following:

\begin{equation} \label{eqn:noisyfl}
    \min_{\theta} {\mathcal{L}}_{\theta} 
    = \sum_{m=1}^{M} \gamma_{m} {\mathcal{L}}_m\left(\theta\right) \textrm{, where } \mathcal{L}_{m}\left(\theta\right)
    = \mathcal{L}_{\theta}\left( I\left(\mathcal{D}_{c}^{m}\right) + \Upphi\left(\mathcal{D}_{n}^{m}\right) \right),
\end{equation}

\noindent where $\mathcal{L}_{m}\left(\theta\right)$ is the supervised loss term of the $m^{th}$ client given model weights $\theta$, $\Upphi\left(\cdot\right)$ is a correction mechanism aiming at reducing the impact of noisy samples of the $m^{th}$ client on the training procedure by either masking or correcting the label $y$ of $\mathcal{D}_{n}^{m}$, and $I\left(\cdot\right)$ is the identity function. With $\gamma_{m}$, we denote the relative impact of the $m^{th}$ client on the generation of the global model $G$. For FedAvg~\citep{fedavg} algorithm, parameter $\gamma_{m}$ is equal to the ratio of client's local data $N_m$ over all training samples, i.e., $\gamma_{m} = \frac{N_m}{N}$.

\subsection{Label Noise Estimation in Federated Setting}\label{sec:noise_estimation}

In the federated setting, label noise is influenced by discrepancies in clients' labeling systems or the expertise of their users. This leads to varying label noise profiles on a per-client basis, where a few `\textit{clean}' clients may exist, holding high-quality labels. To tackle the issue of label noise in FL without introducing unnecessary complexity to clients' computational tasks, we propose simple yet effective approaches for estimating the per-client label noise level. These approaches are designed to accommodate the diverse computational capabilities of devices in a FL system. Specifically, we propose two methods for determining per-client noise level estimation: (i) an embeddings-based discovery, which computes noise from `\textit{noise-tolerant}' embeddings, and (ii) a model's confidence-based approach, where noise is estimated using a scoring function based on the models' outputs or logits. By establishing a per-client noise level, we can efficiently train deep neural networks to improve the performance on a given task, limiting the effect of noisy samples on model's generalizability.

\subsubsection*{\textbf{Embedding-based Discovery of Noisy Labels}} \label{sssec:feature_based}
In supervised learning, corrupted labels can significantly impact the generalization of deep models, leading to poor performance on unseen data~\citep{noises1}. To address this issue, we propose leveraging embeddings from self-supervised pre-trained models, which are trained to learn useful data representations for a variety of tasks~\citep{radford2021learning,shor2022trillsson}, to detect noisy labeled instances in each client's data. By generating embeddings on a per-client basis without relying on any labels, our approach ensures that the extracted embeddings remain robust to the presence of label noise~\citep{knn_paper}.

Formally, we utilize a self-supervised pre-trained model as a feature extractor $g\left(\cdot\right)$ to produce embeddings $e_{i}$ for every input instance $x_{i}{\in}\mathcal{D}_{m}$. With the generation of embeddings, we can then utilize a k-Nearest Neighbor (kNN) approach to identify noisy samples, which corresponds to outliers in the embeddings space (i.e., data points belonging to the same neighborhood with different labels). Specifically, for the neighbourhood of $k$ points surrounding $e_{i}$, we assign a new label using a majority voting mechanism, with random tie-breaking, as:

\begin{equation} \label{eqn:vote_label}
    \begin{aligned}
        y_{i}^{vote}
        =\underset{j\in\left[k\right]}\argmax~\widehat{y_{i}}\left[j\right] 
        = \argmax \sum_{j=1}^{k} \left \{ y_{j} \in \mathcal{D}_{k}: \left| \textit{sorted} \left \{ \left\|g(x_{i})- g(x_{l})\right\|, \forall x_{l} \in \mathcal{D} \right \} \right| < k \right \},
    \end{aligned}
\end{equation}

\noindent where $y_{j}$ is the predicted kNN label for embedding vector $e_{j}$, extracted using the feature extractor $g\left(\cdot\right)$ from an input instance $x_{j}$. One should note that local neighbourhood surrounding embedding $e_{i}$, denoted as $\mathcal{D}_{k}$ in Equation~\ref{eqn:vote_label}, is formulated by computing the euclidean distance of $e_{i}$ with all other embedding's vectors in $\mathcal{D}$. 

To derive a per-client noise level estimation, we divide the number of instances where a mismatch between $y_{i}^{vote}$ and $y_{i}$ occurs over the total number of available samples in each client~\citep{knn_paper}, i.e., $n_{l}^{m} = (\sum_{i=0}^{N_{m}} y_{i}^{vote} \neq y_{i})/N_{m}$. It is important to note that the aforementioned procedure is performed locally on each client during the initialization phase of FL (i.e., beginning of the training). Thus, noise level estimation via embeddings does not introduce additional computation or communication costs to the FL training process.

\subsubsection*{\textbf{Model Confidence as a Proxy for Label Noise}} \label{sssec:score_based}

To provide an alternative to utilizing an external module (i.e.,a pre-trained model) for the detection of {color{mark}noisy labeled instances on a per-client basis, we propose the use of a scoring based method directly applicable to the outputs (or logits) of the neural network.} The intuition behind our approach is to utilize a scoring function to rank input instances, such that a low score indicates a high probability of having a noisy label. Therefore, the critical components required to facilitate this approach include a~\textit{scoring function}, which ranks samples based on model's predictions confidence, and a~\textit{threshold} to distinguish between low-score (noisy labeled) and high-score (correctly labeled) data.

\setlength{\textfloatsep}{0.1cm}
\begin{figure}[!t]
    \centering
    \includegraphics[width=0.4\textwidth]{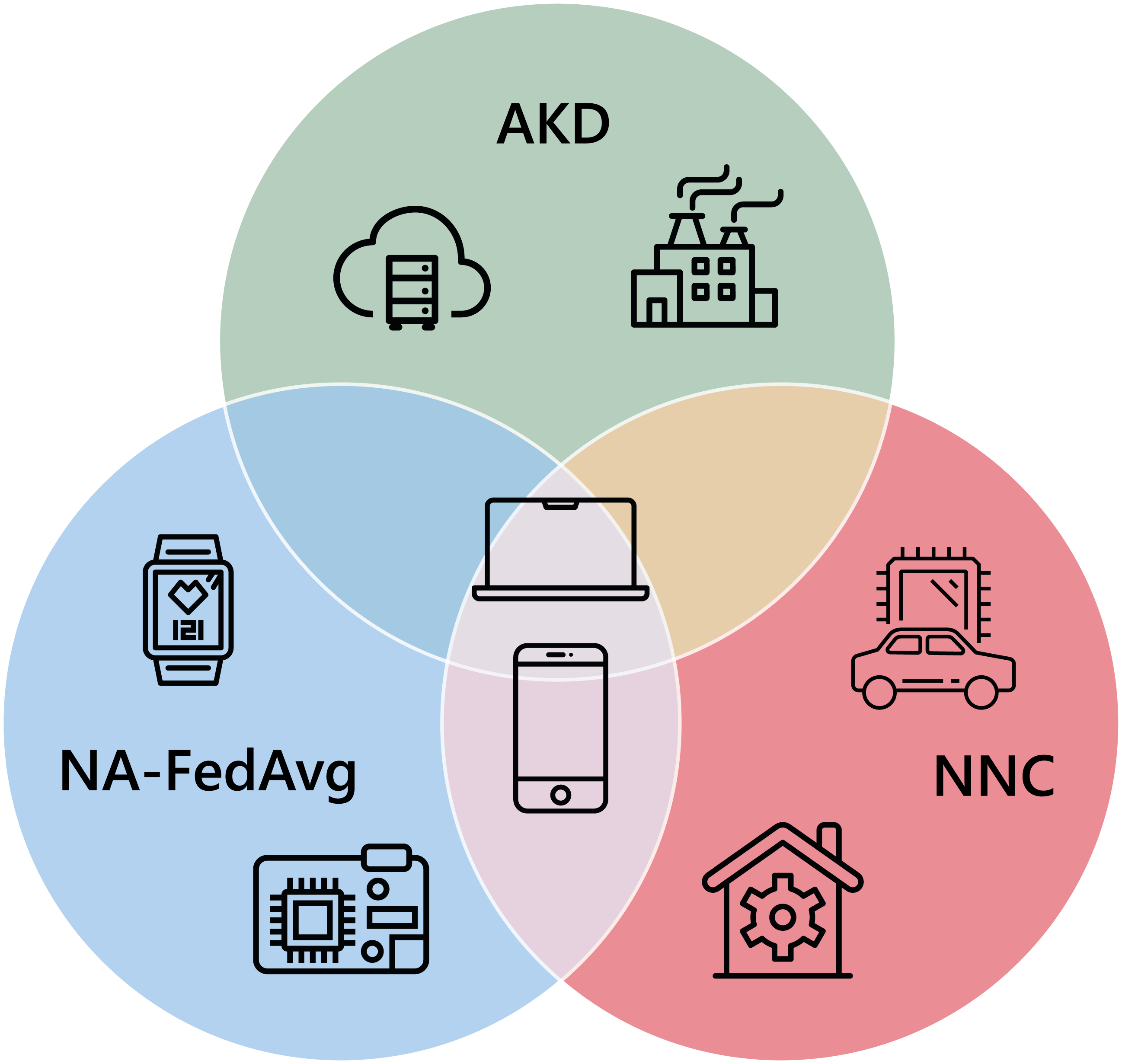}
    \caption{\footnotesize{Illustration of targeted device audience of~\method~approaches. Each approach has different computational resource requirements. \textit{NA-FedAvg} is lightweight, thus well-suited for wearables and edge devices. \textit{NNC} focus on sufficient computational resources for computing embeddings from pre-trained models before training, which are used for label correction, while \textit{AKD} deals with label noise during training by computing an additional loss term, making it suitable for clients with relatively higher computational capabilities.}\label{fig:venn}}
\end{figure}
\setlength{\floatsep}{0.1cm}

To this end, we utilize energy score~\citep{energy_paper} as our~\textit{scoring function} to compute a per-client noise level estimation in a given federated round, denoted as $R_{w}$. While energy score is leveraged for detecting out-of-domain samples, in this work, we propose to use it as a measurement of label uncertainty on clients' data. In contrast to the softmax score, energy score has proven to be less susceptible to modern's neural networks overconfidence issues~\citep{energy_paper,calibration}. Accordingly, we compute the energy score $s$ for an input instance $x$ as the $\mathit{logsumexp}$ ($\mathcal{E}(\cdot)$) operator over the logits $z$, i.e., $s = log(\sum_{i}^{\mathcal{C}} e^{z})$. Specifically, in federated round $R_{w}$, using the newly received aggregated model weights $\theta_{G}^{R_{w}}$, we apply $\mathcal{E}(\cdot)$ operator over each client's dataset to acquire a scoring set $\mathcal{S}_{\theta_{G}^{R{w}}}$, which contains a score for each locally stored sample. In a similar fashion, we utilize the locally trained model weights ($\theta_{m}^{R_{w}}$ - local model weights after completion of local train step in $R_{w}$) to compute a second scoring set, $\mathcal{S}_{\theta_{m}^{R{w}}}$.

The threshold value is computed from the $\nu^{th}$ percentile over $\mathcal{S}_{\theta_{G}^{R{w}}}$, though other statistical measures, e.g., median or mean, can also be used. Specifically, we compute the threshold $\tau_{\nu}$ as:

\begin{equation} \label{eqn:score_threshold}
 \tau_{\nu} = \mathcal{P}_{\nu}\left(\mathcal{S}_{\theta_{G}^{R{w}}}\right) = \underset{\left[\mathcal{S}_{\theta_{G}^{R{w}}}\right]}\argmax~ \left( \frac{\nu}{100} \cdot \left |\mathcal{S}_{\theta_{G}^{R{w}}} \right | \right),
\end{equation}

\noindent where $\left[\mathcal{S}_{\theta^G}\right ]$ and $\left | \mathcal{S}_{\theta^G} \right |$ are the ordered set and cardinality of $\mathcal{S}_{\theta^G}$, respectively. Motivated by the ``\textit{memorization effect}'' in deep networks, where clean data is memorized faster in early training stages~\citep{es1,es2,es3,patrini,4}, an appropriate federated round $R_{w}$ can result in diverse scores between noisy and clean data. Along the same lines, we argue that clients holding high-quality labels have more influential role in the early stages of model aggregation in server-side, aiding in the differentiation between noisy and clean instances on clients' data. Furthermore, by computing a threshold based on the same received global model weights $\theta_G^{R_{w}}$ we provide a ``\textit{common}'' ground for classification, ensuring a clear separation between noisy and clean instances. With both $\mathcal{S}_{\theta_{m}^{R{w}}}$ and $\tau_{\nu}$ computed, we can estimate a per-client noise level by counting the percentage of $m^{th}$ client's local instances that are below the obtained $\tau_{\nu}$, i.e., $n_{l}^{m} = (\sum_{i=0}^{N_m} u_{\tau_{\nu}}(s_{i}))/N_{m}$ with $u_{\tau_{\nu}}(\cdot)$ a ``\textit{$\tau$-shifted}'' Heaviside function that produces $1$ for all inputs above a threshold $\tau$.

Concisely, we estimate the noise level by computing computationally inexpensive scoring sets ($\mathcal{S}_{\theta_{G}^{R_{w}}}$, $\mathcal{S}_{\theta_{m}^{R_{w}}}$) on each client during a single federated round $R_{w}$. This approach is particularly beneficial for devices with low computational and energy resources commonly found in FL. Although the estimated noise level may not be precise compared to the embedding-based method, it is effective in adjusting the importance of clients' updates during server-side model aggregation, as explained in Section~\ref{sec:noisy_fl_train}.

\subsection{Federated Learning under Presence of Label Noise}\label{sec:noisy_fl_train}

The objective of supervised training in presence of label noise is to learn a model, where the effect of such labeled instances on model's performance is minimal. Recent federated approaches often address the issue of label or excessive computational burdens in the  FL process. To tackle this challenge, we propose three distinct approaches, namely NNC, AKD, and NA-FedAvg, each effectively handle label noise at different stages of FL (i.e., initialization, local training steps, and server-side model aggregation) with  varying degrees of compute requirements. Together, these approaches form our framework, termed~\method~, which provide a suitable solutions to label noise in FL for a wide range of devices (low-end to high-end computation devices). At its core,~\method~utilizes label noise estimation techniques (described in Section~\ref{sec:noise_estimation}). In Figure~\ref{fig:venn}, a comprehensive overview of the intended device families, in terms of computational characteristics, associated with each approach is illustrated.

\subsubsection{\textbf{Nearest Neighbor-based Correction (\textit{NNC}):}} \label{sec:nnc}

Input embeddings (i.e., feature extracted from a specific layer of a neural network) computed from a pre-trained model can be exploited to perform label correction. As discussed in Section~\ref{sssec:feature_based}, we estimate the label for each input instance $x$ by ``\textit{looking}'' at the labels of a neighbourhood examples in the embedding space. Thus, apart from predicting a per-client noise estimation using the kNN predictions, we can also perform label correction~\citep{knn_paper}. Specifically, when we detect a label mismatch between the predicted (with kNN) and current label, we consider the predicted label as the true label to be used during the training phase. This way, instead of discarding noisy instances altogether, their labels are modified on a per-client basis at the initialization phase of FL, after which the FL training process continues as usual. Mathematically, the ``\textit{corrected}'' local datasets across all clients in FL can be written as:
 
\begin{equation} \label{eqn:knn_noisyfl_data}
  \mathcal{D}^{m^*} = \left \{ x_{i}, y_{i}^{vote} \right\}_{i=0}^{N_m} = \left \{ x_{i}, \Uppsi \left(g,x_{i} \right) \right\}_{i=0}^{N_m},
\end{equation}

\noindent where $\Uppsi(\cdot)$ is a mapping function that adjusts $y_{i}$ to $y_{i}^{vote}$ (as computed in Equation~\ref{eqn:vote_label}) using the feature extractor $g(\cdot)$. To learn from the ``\textit{corrected}'' datasets across all clients, we apply the cross-entropy loss as:

\begin{equation} \label{eqn:SCE_knn}
    \begin{aligned}
        \mathcal{L}_{\theta}(\mathcal{D}^{m*}) 
        = \mathcal{L}_{CE}\left ( y^{vote}~,~p_{\theta^m}\left(y{\mid} x \right) \right ) 
        = - \frac{1}{N_{m}} \sum\limits_{i=1}^{N_{m}}\sum\limits_{j=1}^{C} {y_{i}^{vote}}^{j} \log (\mathit{f}_{i}^{\theta^{m}}(x_{j})),
    \end{aligned}
\end{equation}

\begin{figure}[t]
    \centering
    \includegraphics[width=0.85\textwidth]{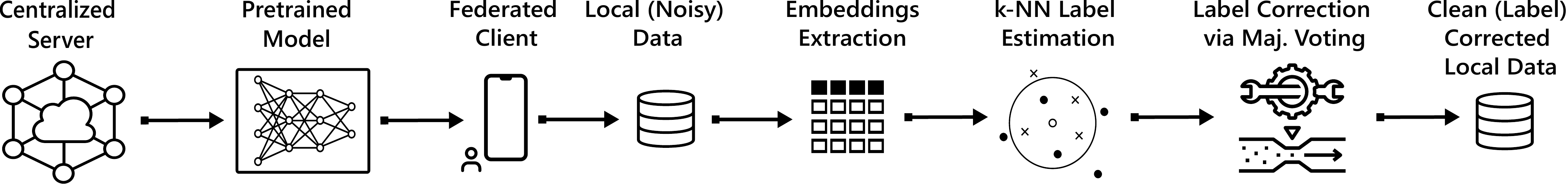}
    \caption{\footnotesize{Illustration of \textit{NNC} for identifying and correcting noisy labeled instances via embeddings. The process is depicted for one federated client for the sake of simplicity.}}
    \label{fig:nnc_arch}
\end{figure}

It is important to highlight that the quality of the extracted embeddings is crucial for the effectiveness of \textit{NNC} in identifying and correcting noisy instances. However, we do not see it as a major problem as several self-supervised pre-training approaches~\citep{radford2021learning, saeed2021contrastive, shor2022trillsson} provide useful embeddings for broad spectrum of tasks. Alternatively, the label correction process in \textit{NNC} is not affected to the same degree by the actual label noise present at each client, since the embeddings are extracted without any labels. In addition, clients are required to hold a pre-trained model only for a single forward-pass during the initialization phase of FL. Further details and an overview of our proposed NNC federated approach for learning models under the presence of label noisy can be found in Algorithm~\ref{alg:method}, highlighted in red color.

\subsubsection{\textbf{Adaptive Knowledge Distillation (\textit{AKD}):}} \label{sec:akd}

\begin{figure}[t]
    \centering
    \includegraphics[width=0.55\textwidth]{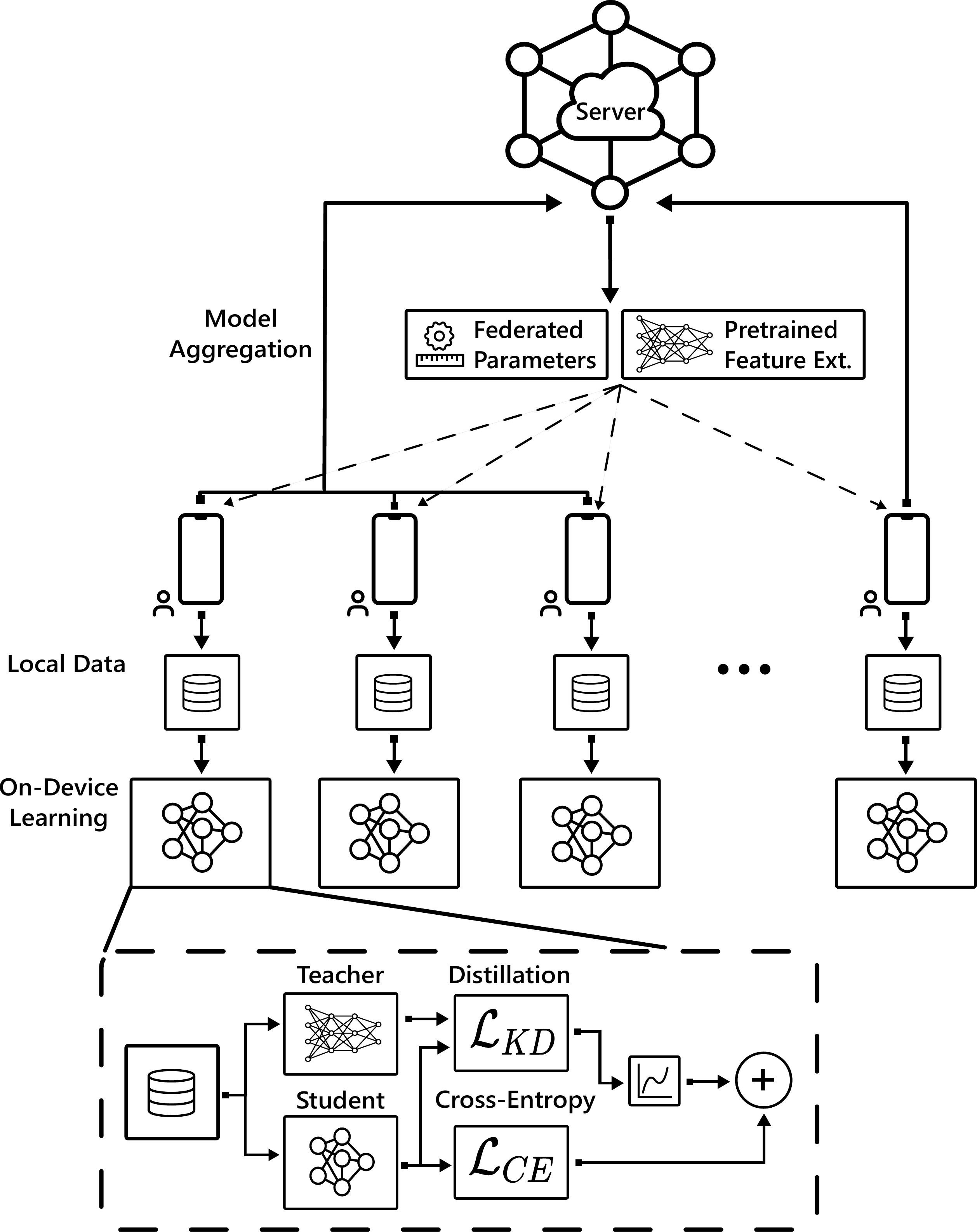}
    \caption{\footnotesize{Illustration of our \textit{AKD} approach for training federated models under the presence of noisy labels with "\textit{adaptive}" Knowledge Distillation (see Section~\ref{sec:akd}). Embeddings are extracted from a pre-trained model, as an extra source of supervision for local models, while KD is activated on a per client-basis in case label noise is detected.}\label{fig:akd_arch}}
\end{figure}

Rather than directly altering the labels, we can implicitly guide the model to avoid ``\textit{memorization}'' of noisy labeled samples during the FL training phase through means of an additional loss term. Specifically, in addition to the standard cross-entropy loss, we propose to use a knowledge distillation loss~\citep{hinton, fitnets} that requires each client to also learn to mimic embeddings or output of a teacher model. In contrary to the NCC, in which we proposed a label correction process for training models under the presence of noisy labeled instances, with \textit{AKD}, we utilize soft-labels or embeddings from a teacher model as ground-truth. 

To this end, we propose to utilize embeddings computed with a pre-trained model or the server-side aggregated model's outputs from each round as a source of supervision. Specifically, after the noise level being estimated on a per-client basis, as discussed in Section~\ref{sec:noise_estimation}, we incorporate a distillation loss term to clients with detected label noise, while permitting any client with well-annotated data to directly train on their locally stored data. Thus, we present an ``\textit{adaptive}'' knowledge distillation (\textit{AKD}) during FL training process, where clients with noisy instances are guided with additional supervision to learn noise-robust models. To learn from the datasets of all clients, $\mathcal{D}^{m}$, we apply the cross-entropy loss and an knowledge distillation loss as: 

\begin{equation} \label{eqn:SCE_akd}
    \resizebox{0.91\hsize}{!}{%
        $
        \begin{aligned}
            \mathcal{L}_{\theta}(\mathcal{D}^{m}) &= 
            \mathcal{L}_{CE}\left (y,p_{\theta^m}\left(y\mid x \right) \right) + \beta \cdot u_{\epsilon}(n_l^{m}) \cdot \mathcal{L}_{KD}\left(e, p_{\theta^{m}}\left(e|x\right)\right ) \\   
            & \textrm{ where } , \mathcal{L}_{KD} =
            \begin{cases}
                \mathcal{L}_{MAE} = \frac{1}{N_{m}} \sum\limits_{i=1}^{N_{m}}  \left | e_{i} - \tilde{e_{i}} \right | & \text{,~if $e$ are embeddings of $g(\cdot)$} \\
                \mathcal{L}_{KL} =  T^{2} \sum\limits_{j=1}^{C} p_{\theta^m}^{T} log \frac{p_{\theta^m}^{T}}{p_{\theta^G}^{T}} & \text{,~if $e$ are ``\textit{temperature-scaled}'' logits of $G$} \\ 
            \end{cases}
        \end{aligned}
        $
        }
\end{equation}

\noindent Here, $\mathcal{L}_{CE}(\cdot)$ indicates the standard cross-entropy loss, $\mathcal{L}_{MAE}(\cdot)$ corresponds to the mean absolute error loss between the embeddings computed from a feature extractor $g(\cdot)$ and the local model of the $m^{th}$ client (noted as $\tilde{e_{i}}$), and $\mathcal{L}_{KL}(\cdot)$ refers to the Kullback-Leibler divergence loss~\citep{hinton} between the ``\textit{temperature-scaled}'' logits of $m^{th}$ client's local model and the server-side aggregated model $G$ of the current federated round. With ``\textit{$\epsilon$-shifted}'' Heaviside function, $u_{\epsilon}$, we introduce $\mathcal{L}_{KD}$ loss to clients, whose estimated label noise exceeds $\epsilon$, while we add the scalar $\beta$ to control the contribution of $\mathcal{L}_{KD}$ on model optimization. We fix $\beta$=10 and $T$=2, which we found to be working well during our initial exploration. Further details and an overview of our proposed \textit{AKD} federated approach can be found in Algorithm~\ref{alg:method}, highlighted with green color.

\subsubsection{\textbf{Noise-aware Federated Averaging (\textit{NA-FedAvg}):}}\label{sec:na_fedavg}

\setlength{\textfloatsep}{0.1cm}
\begin{figure}[t]
    \centering
    \includegraphics[width=0.55\textwidth]{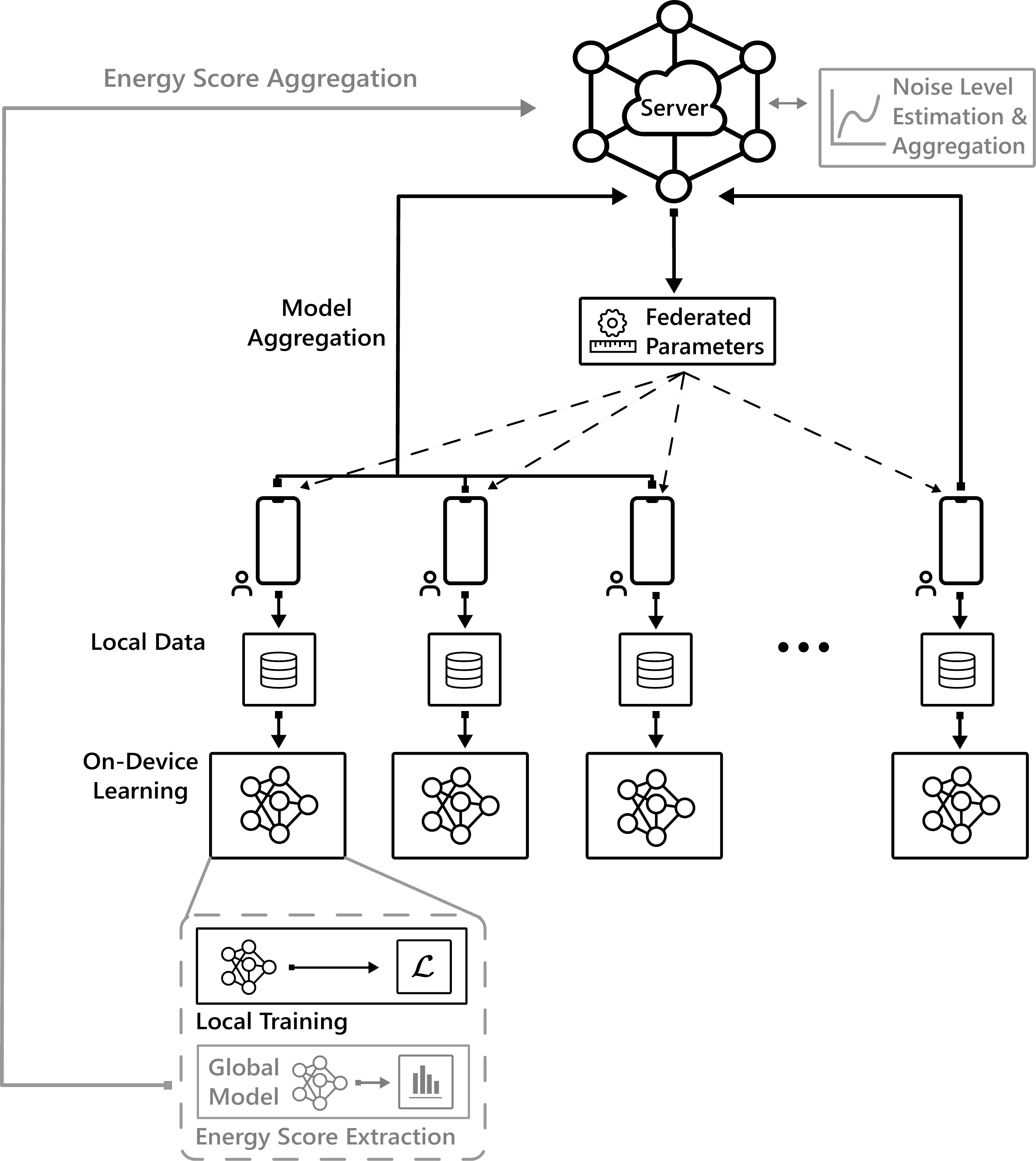}
    \caption{\footnotesize{Illustration of NA-FedAvg method for mitigating the effect of label noise on model's generalizability. In addition to the standard FedAvg process, a noise estimation process is introduced once, indicated with grey colors. In this round, energy scores are computed and communicated to server. Afterwards, the global model and a noise estimation per client is calculated on server. For the remaining FL training rounds, a typical local train step is performed, while server performs a noise-aware weighted model aggregation, incorporating clients' estimated noise level.} \label{fig:na_fedavg_arch}}
\end{figure}
\setlength{\floatsep}{0.1cm}

As an alternative to previously proposed approaches for dealing with noisy labels, we aim to directly address the impact of noisy labels during a later stage of the training process. Specifically, during the server-side aggregation process of \textit{FedAvg}~\citep{fedavg}, we propose utilizing the estimated noise level of each client to perform a ``\textit{noise-aware}'' \textit{FedAvg} (\textit{NA-FedAvg}) aggregation step by modifying parameter $\gamma_{m}$ in Equation~\ref{eqn:noisyfl} to consider both the number of samples and the label noise level of clients. In other words, clients with few noisy instances have a greater contribution to the FL process compared to the ones with a large amount of label noise. While straightforward, weights re-scaling techniques have been beneficial in mitigating the impact of noisy labels on the global model $G$; yet relying on additional clean data to provide a computationally inexpensive label noise estimation in FL~\citep{8,9,10}. In this work, to minimize the computational overhead for clients in \textit{NA-FedAvg}, we directly estimate the noise level from the model's predictions confidence, as discussed in Section~\ref{sssec:score_based}. This enables \textit{NA-FedAvg} to be suitable for clients with limited computational resources. Clients only compute computationally inexpensive scores for each locally stored sample, while the majority of the work is off-loaded to the server, which computes a noise level estimation for each client.

\setlength{\textfloatsep}{0.1cm}
\begin{algorithm}[t]
    \tiny
    \caption{\footnotesize{~\method: Federated learning with Label Noise. We develop three distinct approaches to deal with label noise in learning models from decentralized data, whereas FedAvg~\citep{fedavg} is the base algorithm and it is to indicate the key contributions of our proposed approaches. In the algorithm, scalar $R_{w}$ indicate the activation round for NA-FedAvg algorithm and $\eta$ is the learning rate.(\colorbox{_yellow}{FedAvg}, \colorbox{_red}{NNC}, \colorbox{_green}{AKD}, \colorbox{_blue}{NA-FedAvg}})\label{alg:method}}
    
    \begin{algorithmic}[1]
        \State Server initialization of model $G$ with model weights $\theta_{0}^{G}$, $n_{l}^{m}$=$0$, $\forall m \in M$
        \For{ each client $m \in M$ \textbf{ in parallel}}
            \For{ $\left ( x_{i},y_{i} \right ) \in \mathcal{D}^m$}
                \State \colorbox{_red}{$\tilde{y_{i}} \gets \Upphi \left (g, x_{i} \right )$}
                \State \colorbox{_green}{$e_{i} \gets g\left (x_{i} \right )$}
            \EndFor
            \State \colorbox{_green}{$n_{l}^{m} = \frac{ \sum_{i=0}^{N_{m}} y_{i}^{vote} \neq y_{i} }{N_{m}} $}
        \EndFor
        
    \For{ $r=1,\dots,R$ }
        \State Randomly select $M$ clients to participate in round $i$
            \For{ each client $m \in M$ \textbf{ in parallel}}
                \State $\theta_{r}^{m} \gets \theta_{r}^{G}$
                \State $\theta_{r+1}^{m}$,  \colorbox{_blue}{($S_{\theta_{r}^{G}}^m,S_{\theta_{r+1}^{m}}^m$)} $\gets$ ClientUpdate($\theta_{r}^{m}$,$r$,$n_l^m$)
            \EndFor
            \IIf{$r = R_{w}$} \colorbox{_blue}{$n_l^m = \frac{\sum\nolimits_{i=0}^{N_m} u_{\tau_{\nu}} \left(s_{i,\theta^{m}}^{m}\right)}{N_m}$} \EndIIf
            \State \colorbox{_yellow}{$\theta_{r+1}^{G} \gets \sum\nolimits_{m=1}^{M} \frac{N_m}{N} \theta_{r+1}^m$}
            \State \colorbox{_blue}{$\theta_{r+1}^{G} \gets \sum\nolimits_{m=1}^{M} \left (1-n_{l}^{m} \right) \cdot \frac{N_m}{N} \theta_{r+1}^m$}
    \EndFor

    \Procedure{ClientUpdate}{$\theta$, $r$, $n_{l}$}
        \For{epoch $e=1,2,\dots,E$}
            \For{ batch $b \in \mathcal{D}^{k}$}
                \State \colorbox{_red}{$\acute{\theta} \gets \theta - \eta \nabla_{\theta} \left( \mathcal{L}_{CE} \left(y^{vote}, p_{\theta}\left(y|x_{b}\right) \right) \right)$}
                \State \colorbox{_green}{$\acute{\theta} \gets \theta - \eta \nabla_{\theta} \left( \mathcal{L}_{CE} \left(y, p_{\theta}\left(y|x_{b}\right) \right) + \beta \cdot u_{\epsilon}(n_l) \cdot \mathcal{L}_{KL} \left(e_b, p_{\theta}\left(y|x_{b}\right) \right) \right)$}    
                \State \colorbox{_yellow}{$\acute{\theta} \gets \theta - \eta \nabla_{\theta} \left( \mathcal{L}_{CE} \left(y, p_{\theta}\left(y|x_{b}\right) \right) \right)$}
                \IIf{$r = R_{w}$} \colorbox{_blue}{$ s_{b,\acute{\theta}} \gets \mathcal{E}\left(x_{b},p_{\acute{\theta}}\right) ~,~ s_{b,\theta} \gets \mathcal{E}\left(x_{b},p_{\theta}\right)$} \EndIIf
           \EndFor
        \EndFor
        \State \Return $\acute{\theta}$, \colorbox{_blue}{($S_{\theta}$,$S_{\acute{\theta}}$)}
    \EndProcedure
    \end{algorithmic}
\end{algorithm}
\setlength{\floatsep}{0.1cm}

Concisely, \textit{NA-FedAvg} begins with standard FL process up to a certain federated round $R_{w}$, which is considered a hyperparameter of \textit{NA-FedAvg}. In round $R_{w}$, after receiving globally aggregated weights $\theta_{G}^{R_{w}}$, each client performs a local training step on the global model weights and now possesses a local model with weights $\theta_{m}^{R_{w}}$. Using these two local and global models, and only for round $R_{w}$, the scoring sets $\mathcal{S}_{\theta_{G}}^{R_{w}}$ and $\mathcal{S}_{\theta_{m}}^{R_{w}}$ are computed on a per-client basis and communicated back to the central server, together with each model's parameters $\theta_{R_{c}}^{m}$. Using the acquired scoring sets, we compute a per-client noise estimation at the server-side, as discussed Section~\ref{sssec:score_based}, to introduce the ``\textit{noise-aware}'' model aggregation by setting $\gamma_{m}= (1-n_{l}^{m})\cdot\frac{N_m}{N}$ in Equation~\ref{eqn:noisyfl}, while the remaining FL training process continues as usual. Further details and an overview of our ``\textit{Noise-Aware}'' FedAvg approach can be found in Algorithm~\ref{alg:method}, highlighted in blue color.

\section{Experiments}\label{experiments}
In this section, we describe our extensive performance evaluation for~\method. We use various publicly available datasets to determine efficacy of our approaches in learning generalizable models under a variety of federated leaning and label noise settings. Firstly, the utilized datasets are presented, followed by a detailed description of the neural network architectures and FL framework used in our experiments. Next, we present our evaluation strategy, including all considered federated parameters and baselines used to evaluate/compare against our methods. Finally, we provide our finding about performance of~\method~across a wide range of label noise scenarios.

\subsection{Datasets}\label{ssec:datasets}

We use publicly available datasets for performance evaluation on a range of classification tasks from both the vision and audio domains. For all datasets, we use standard training/test splits for comparability purposes, as provided with the original datasets. From the vision domain, we use the CIFAR-10~\citep{cifar10} and FashionMNIST~\citep{fmnist} and EuroSAT datasets, where the tasks of interests are object detection, clothes classification and landmark categorization, respectively. By utilizing these datasets, we facilitate the ease of benchmarking for future research. In addition, we perform experiments with the PathMNIST\citep{pathmnist} dataset from the medical imaging. By doing this, we investigate the performance of~\method~in a domain, where noisy labels may occur due to incorrect diagnosis. From the audio domain, we use SpeechCommands (v2) dataset~\citep{spcm}, where the learning objective is to detect when a particular keyword is spoken out of a set of twelve target classes. Apart from these datasets, we extend our evaluation to a real-world, human annotated version of popular CIFAR-10/100~\citep{cifarn} datasets, namely CIFAR-10N/100N, where label noise presents varied (or biased in some manner) patterns based on users' preferences.

\begin{table}[t]
    \scriptsize \centering
    \caption{\footnotesize{Details from our experimental setup. In (a) details of the considered datasets are presented, while (b) describes our primary parameters used in our experiments.}\label{tab:setup}}
    \begin{minipage}[t]{0.47\hsize} \centering \resizebox{1.0\hsize}{!}{%
    \begin{tabular}{lcc}
        \toprule
        \textbf{Dataset} & \textbf{Task} & \textbf{Classes} \\ \midrule
        CIFAR-10~\citep{cifar10} & Object detection & 10 \\
        FashionMNIST~\citep{fmnist} & Clothing classification & 10 \\
        PathMNIST~\citep{pathmnist} & Pathology reporting  & 9 \\ 
        EuroSAT~\citep{eurosat} & Landmark classification  & 10 \\ 
        SpeechCommands~\citep{spcm} & Audio keyword spotting & 12 \\
        \midrule
        CIFAR-10N~\citep{cifarn} & \multirow{2}{*}{Object detection} & 10 \\
        CIFAR-100N~\citep{cifarn} &  & 100 \\
        \bottomrule
    \end{tabular}%
    }
    \vskip.8\baselineskip
    (a) Details of the datasets used in our experiments 
    \end{minipage}%
    \hfill
    \begin{minipage}[t]{0.47\hsize} \centering \resizebox{1.0\hsize}{!}{%
        \begin{tabular}{@{}lcc@{}}
            \toprule
            \multicolumn{1}{c}{\textbf{Name}} & \textbf{Parameter} & \textbf{Range} \\ 
            \midrule[0.5pt] 
            Number of Clients                               & $M$       & 30 \\
            Number of Federated Rounds                      & $R$       & 1\textemdash 200 \\
            Number of Local Train Steps                     & $E$       & 1 \\
            Clients' Participation Rate                     & $q$       & 80\%  \\
            Noise Level                                     & $n_l$     & 0\textemdash 100\%  \\
            Noise Sparsity                                  & $n_s$     & 0\textemdash 100\% \\
            Percentage of Noisy Clients                     & $F$       & 0\textemdash 100\% \\
            Data Distribution Variance across Clients       & $\sigma$  & 25\% \\
            \bottomrule
        \end{tabular}%
        }
        \vskip.8\baselineskip 
        (b) Primary experiment parameters
    \end{minipage}
\end{table}

For all image classification tasks, we perform standard augmentations, such as random flipping and cropping, followed by Cutout~\citep{cutout} transformation. For the SpeechCommands audio dataset, we extract log-Mel spectrograms from raw waveforms as our model input. We compute this by applying a short-time Fourier transform on the one-second audio segment with a window size of $25$ \textit{ms} and a hop size equal to $10$ \textit{ms} to extract $64$ Mel-spaced frequency bins for each window. In order to make an accurate prediction on an audio clip, we average over the model predictions of non-overlapping segments of an entire audio clip.

\subsection{Implementation Details and Evaluation Strategy} \label{sec:implementation}

\subsubsection*{\textbf{Models and Optimization}}\label{ssec:model} For our image classification tasks, we choose a ResNet-20~\citep{resnet} model architecture due to its relatively compact model size, which makes it ideal for on-device learning, where devices have medium to low computational resources. Here, we utilize an SGD optimizer with a learning rate of $0.1$ and momentum of $0.9$ for both CIFAR-10 and FashionMNIST, while the Adam optimizer with the default learning rate of $0.001$ was used for the PathMNIST and EuroSAT.

For the audio domain, the network architecture of our global model is inspired by~\citep{Model} for mobile devices. Our convolutional neural network architecture consists of four blocks. In each block, we perform two separate convolutions: one on the temporal and another one on the frequency dimension, outputs of which we concatenate afterwards in order to perform a joint $1 \times 1$ convolution. Using this scheme, the model can capture fine-grained features from each dimension and discover high-level features from their shared output. Furthermore, we apply L$2$ regularization with a rate of $0.0001$ in each convolution layer and group normalization after each layer. Between model blocks, we utilize max-pooling to reduce the time-frequency dimensions by a factor of $2$ and use a spatial dropout rate of $0.1$ to avoid over-fitting. We apply ReLU as a non-linear activation function and use Adam optimizer with the default learning rate of $0.001$ to minimize the loss function.

For our methods as discussed in Sections~\ref{sssec:feature_based} and~\ref{sec:nnc}, which relies on embeddings, we use Off-the-self pretrained models trained on large-scale datasets in an unsupervised manner. For vision models, we use ViT-B/32 from CLIP~\citep{radford2021learning} and for audio we leverage TRILLsson (v3, EfficientNetv2-B3)~\citep{shor2022trillsson}, which has a same audio front-end as we used for our audio model. These publicly available models can be downloaded directly on client devices and we run them once (i.e., forward-pass only) to compute embeddings from client's local storage.

\subsubsection*{\textbf{Federated Environment}}\label{ssec:flower} To simulate a federated environment, we use the Flower framework~\citep{Flower} and utilize FedAvg~\citep{fedavg} as the optimization algorithm to construct the global model from clients' local updates. Additionally, a number of primary parameters, listed in Table~\ref{tab:setup}b, were selected to control the federated setting in our experiments. In all~\method~experiments, where noise level estimation is computed based on a model's confidence, we use a fixed scoring threshold percentile $\nu$ to $75\%$, while for embedding-based noise discovery the neighborhood size in kNN to $100$, which we found to be working well during our initial exploration. Further, we set the clients' participation rate in each federated round ($q$) to be equal to $80$\%. It is important to note that we employ uniform random sampling for the clients' selection strategy, as other approaches for adequate clients election are outside the scope of our work.

For the data distribution process in our federated experiments, we randomly partitioned the datasets across the available clients in a non-overlapping fashion, controlling the amount of data across clients through parameter $\sigma$. With $\sigma$ set to $25$\% and a random partitioning of data among clients, the resulting data distribution across clients is intentionally imbalanced. This type of data distribution is common across clients' data in federated setting~\citep{OpenChallenges}, where the variation in the number of samples per client is influenced by the characteristics of the respective users. To generate label noise in the datasets, we constructed a noise matrix $Q_{y{\mid}y^{*}}$ based on parameters $n_{l}$ and $n_{s}$, similar to~\citep{cl}. While in centralized settings a single noise matrix was considered, in FL we constructed a unique noise matrix per client, thus introducing distinct noise profiles across clients. We note that the noise injection process has been performed after the data partioning process. With parameter $F$, we controlled the number of ``\textit{noisy}'' clients, i.e., clients' holding noisy labeled instances in their datasets. Lastly, for an accurate comparison between our experiments, we manage any randomness during data partitioning, label noise injection and training procedures by using a seed alongside the parameters as presented in Table~\ref{tab:setup}.

\subsubsection*{\textbf{Benchmark Baselines}}\label{ssec:baselines}

As the effect of label noise remains mostly unexplored in federated setting, we perform a thorough evaluation across a wide range of existing techniques used in centralized setting to handle noisy labels. From the perspective of regularization techniques, we consider Label Smoothing~\citep{ls}, which ``\textit{softens}'' the labels by taking a weighted average of the hard targets and the uniform distribution over labels. Such ``\textit{smoothing}'' of labels aims to accounts for the fact that datasets could contain mislabeled instances, thus maximizing the likelihood of $p\left(y{,}x\right)$ directly can be harmful. In our experiments, we adjusted the Label Smoothing rate, $\alpha$, to be equal to $0.2$~\citep{ls_rate}. Additionally, we conduct experiments, where we use Bi-Tempered~\citep{bTL} loss instead of standard cross-entropy. This loss replaces softmax function with a high temperature generalization and uses a low temperature logarithm. In this way, Bi-Tempered loss is able to construct a decision boundary less susceptible to noisy labels and learn from outliers (which are considered noisy labeled instances) present in the data. Furthermore, from a noisy instances detection perspective directly from data, we consider Confidence Learning (CL)~\citep{cl}. It prunes noisy instances from a dataset based on probabilistic thresholds from a neural network's predictions. To compute these probabilistic thresholds, we utilize the globally unified model $G$ during the initialization phase of FL, which we train locally for a fixed number of epochs ($E$=$20$) on all clients. Once the data pruning was performed using these thresholds, we discard the trained model and resume the FL training process to train a model from scratch. 

From existing FL approaches, we utilized FedCorr~\citep{4}, a multi-stage label noise correction approach based on Gaussian Mixture Models (GMM) and pseudo-labeling, in our federated learning experiments. FedCorr consists of a pre-processing stage where a subset of clients participate in each iteration (all clients are participating per iteration in random subsets) to detect noisy labeled instances using a GMM based on samples LID~\citep{lid} values. The label correction phase involves training the model with only the relatively "clean" clients (i.e., clients with low label noise) and generating pseudo-labels for the identified noisy instances. Subsequently, a standard federated learning (\textit{FedAvg}) process is performed on all clients. We followed the original FedCorr methodology, conducting 5 iterations with a 1\% participation ratio for a total of 150 rounds. For fine-tuning and standard federated learning, we conducted 95 and 100 rounds, respectively, to align with other approaches used in our experiments. As MixUp~\citep{mixup} augmentation is utilized in FedCorr, we did not consider FedCorr for our audio classification tasks. Apart from these methods, we perform preliminary experiments under standard centralized and federated setting with clean data. With the centralized experiments, we aim to establish an upper bound of performance for our FL approach, while the FL experiments serve a baseline to evaluate the performance improvement we can obtain with~\method, when noisy labels are present. Lastly, for a rigorous evaluation, we perform three distinct trials (i.e., running an entire federated experiment) in each setting, and the average accuracy over three runs is reported across the results of Section~\ref{ssec:results}.

\subsection{Results} \label{ssec:results}

In this subsection, we discuss our findings on the effectiveness of~\method~to deal with label noise in FL. First, we show the efficacy of our noise detection mechanisms, Next, we study the performance of~\method~across a wide range of classification tasks. Finally, we explore how our approaches handle different model architectures, distinct noise profiles, and real-life label noise settings.

\subsubsection{\textbf{Mechanisms for label noise detection in FL}} \label{sssec:noisy_samples}\hfill
\setlength{\textfloatsep}{0.1cm}
\begin{figure}[!t]
    \centering
    \subfigure[$n_l$=40\% and $n_s$=0\%.]{\label{fig:det1} \includegraphics[width=.23\textwidth]{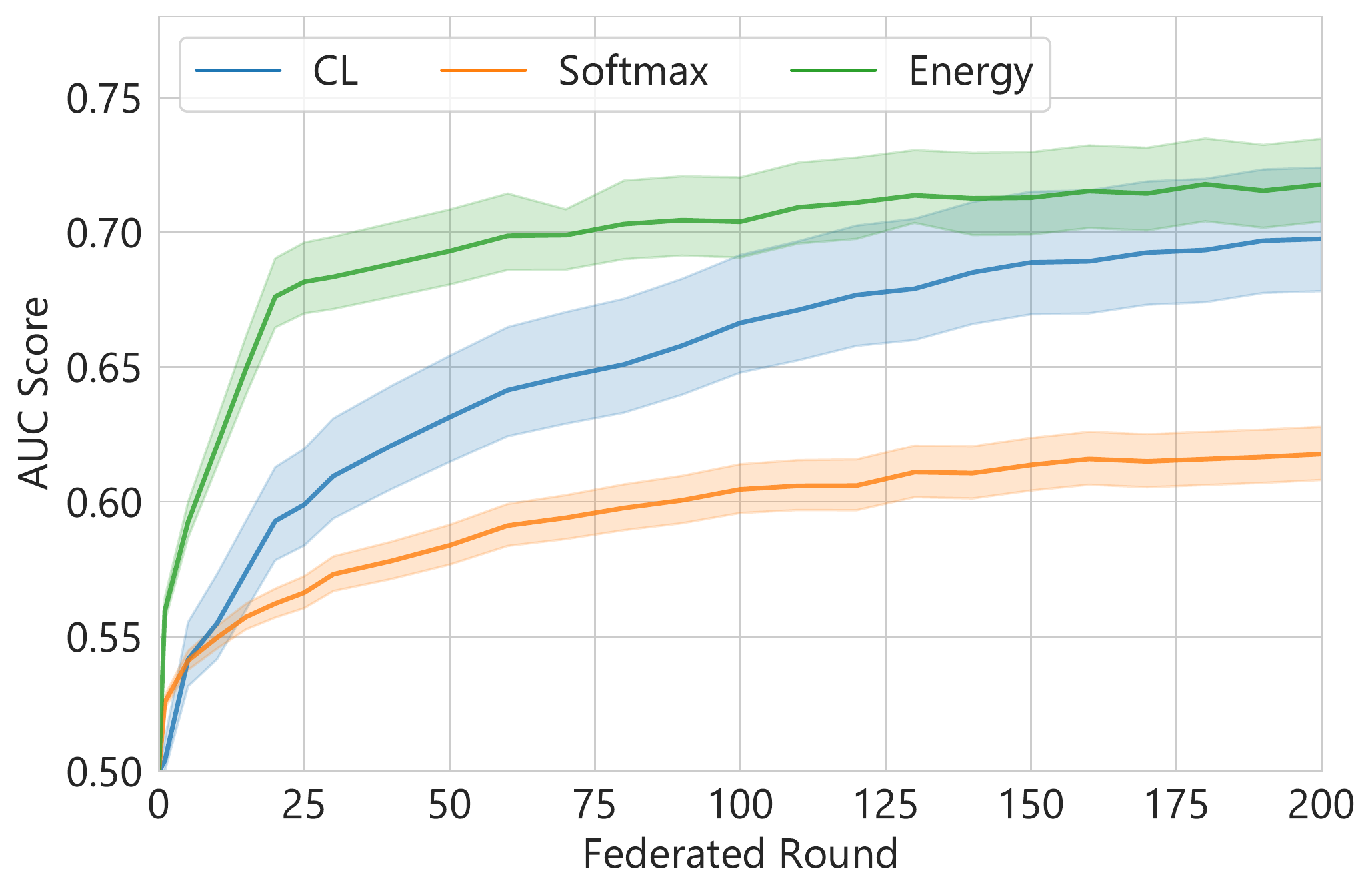}}
    \subfigure[$n_l$=40\% and $n_s$=40\%.]{\label{fig:det2} \includegraphics[width=.23\textwidth]{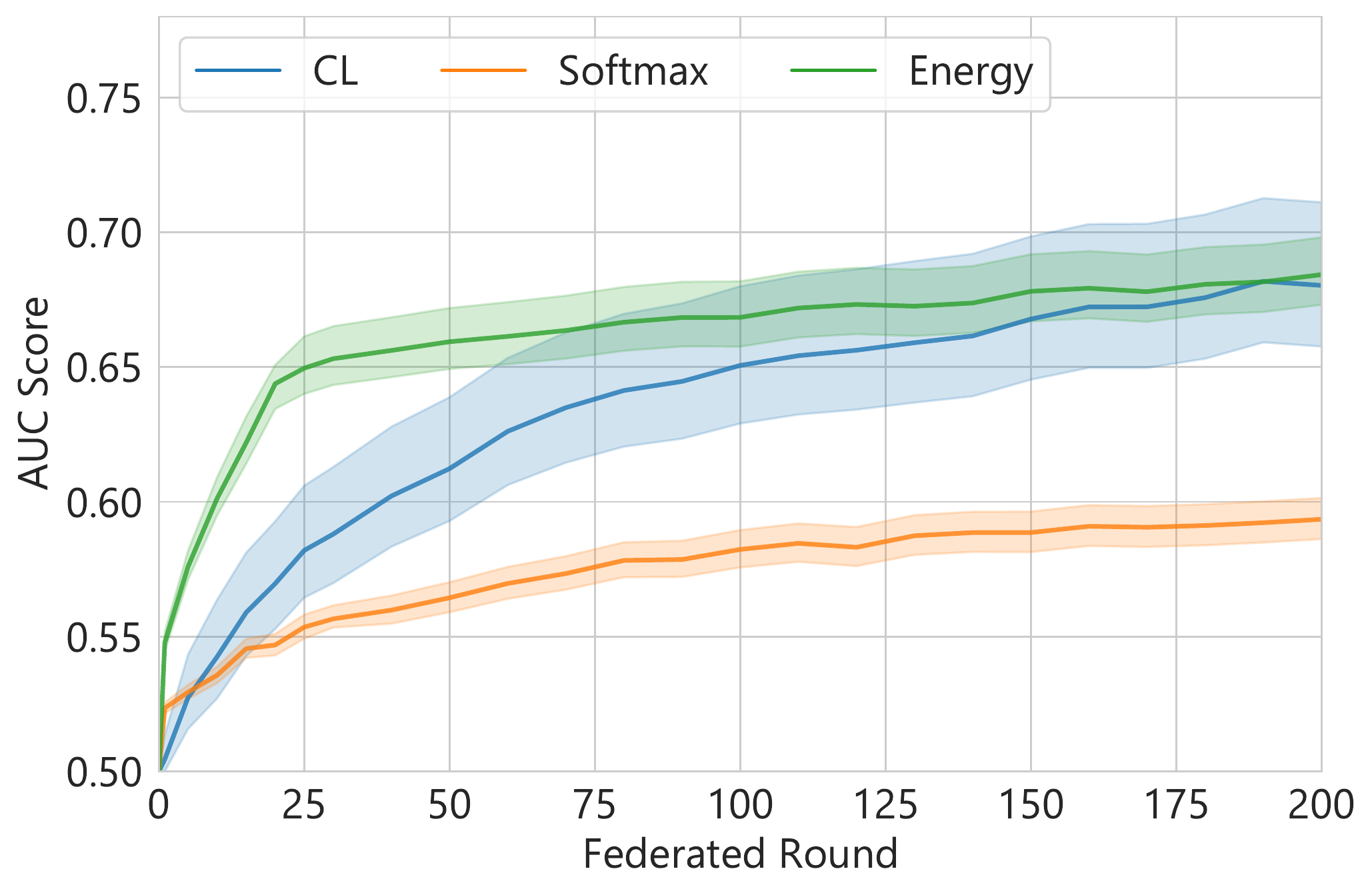}}
    \subfigure[$n_l$=70\% and $n_s$=0\%.]{\label{fig:det3} \includegraphics[width=.23\textwidth]{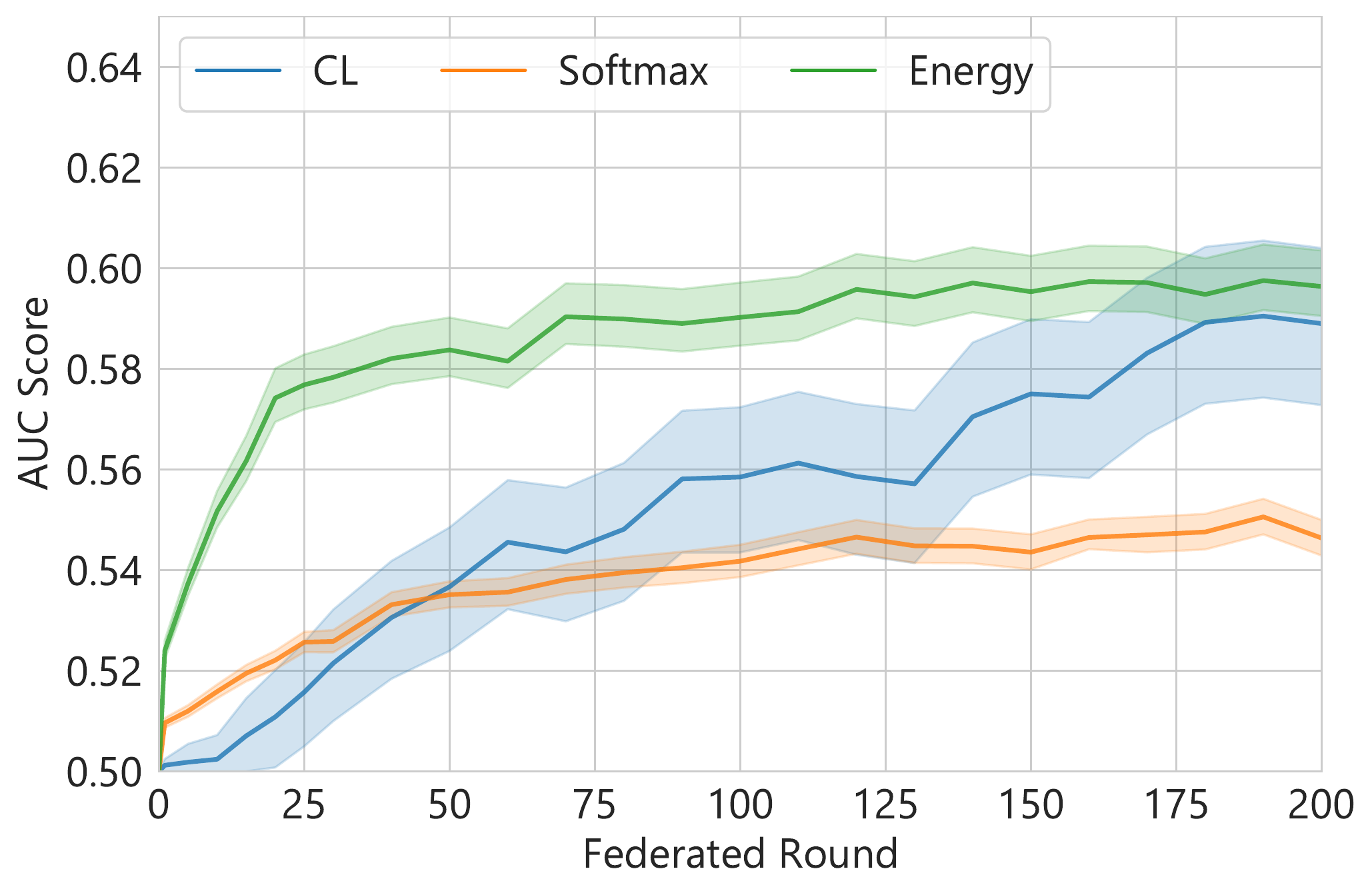}}
    \subfigure[$n_l$=70\% and $n_s$=70\%.]{\label{fig:det4} \includegraphics[width=.23\textwidth]{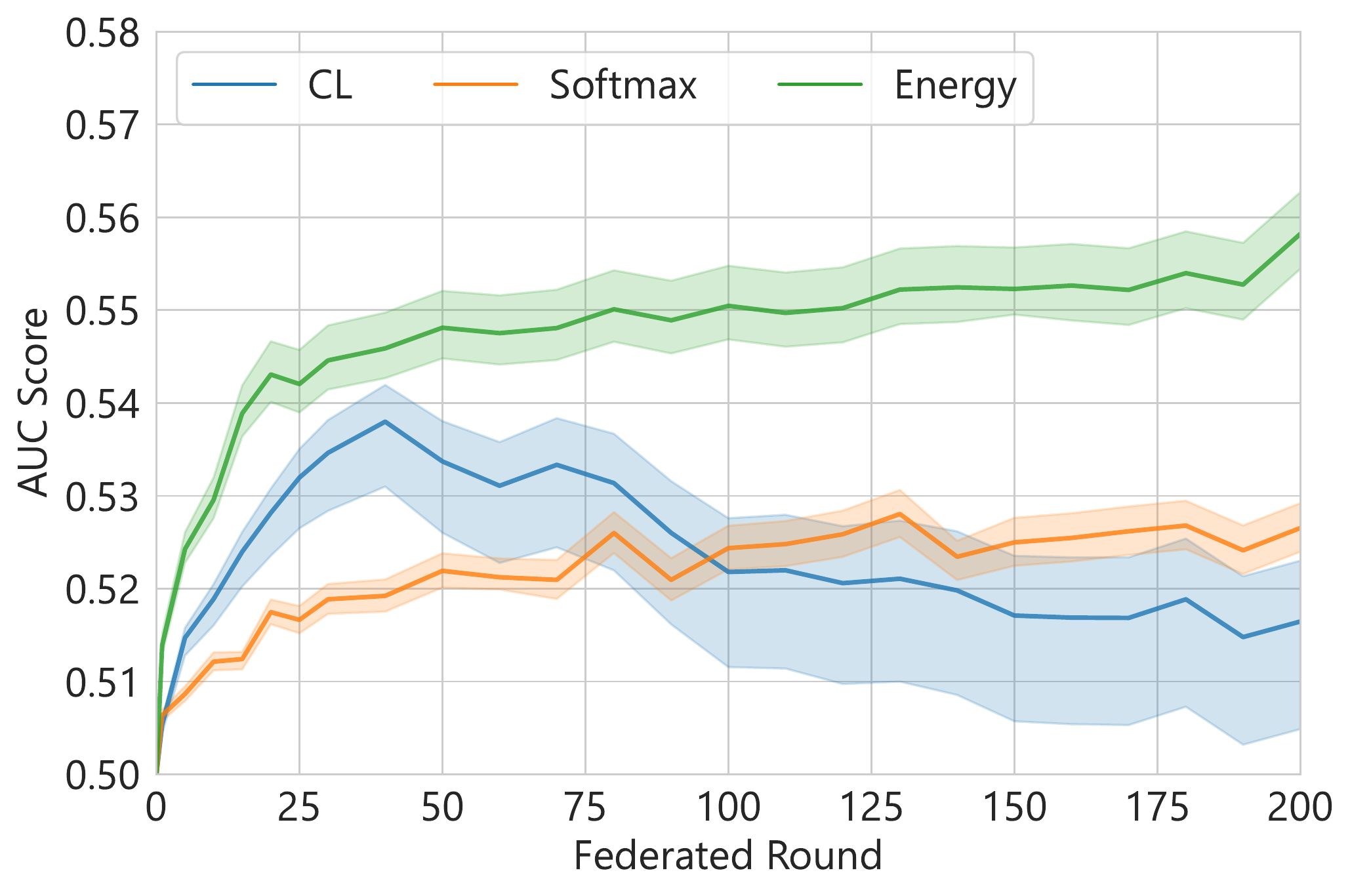}}
    \vspace*{-0.4cm}
    \caption{\footnotesize{Performance evaluation of model confidence based mechanisms for detection of noisy labeled instances. CL refer to Confidence Learning~\citep{cl}, while Softmax and Energy~\cite{energy_paper} correspond our approach, when the corresponding scoring function is being utilized. The mean AUC score across clients in different federated rounds is reported on CIFAR-10, while variance is indicated from the shaded line. Federated parameters are set to $R$=200, $M$=30, $F$=80\%, $E$=1, $q$=80\%, and $\sigma$=25\%.}\label{fig:noise_estimation}}
\end{figure}
\setlength{\floatsep}{0.1cm}

The estimation of noise level on a per-client basis is a fundamental block of our~\method~framework. By exploiting this information, we can compose learning schemes to mitigate the effect of label noise during training. For this purpose, we first perform experiments, where we evaluate the performance of our proposed label noise detection mechanisms, namely \textit{embedding-based} and \textit{model's confidence-based}, on CIFAR-10 and SpeechCommands for different noise profiles. Additionally, we conduct experiments, where, model's predictions (i.e., softmax scores), instead of energy scores, were used to provide a noise level estimation. From our considered baselines, we used Confidence Learning~\citep{cl} data pruning process as an approximation to noise level, where the noise level correspond to the percentage of pruned data, while we also include FedCorr~\citep{4} estimated noise label across clients. We perform experiments with identical data partitioning and noise injection processes across methods to ensure an unbiased evaluation.

While the embedding-based discovery of noisy labels does not depend on the clients' models, our \textit{energy-based} approach require proper tuning of $R_{w}$, as discussed in Section~\ref{sec:noisy_fl_train}. To this end, we performed initial experiments to derive a suitable $R_w$, where noisy labeled instances can be effectively detected across clients. Figure~\ref{fig:noise_estimation} provide obtained AUC score on detection of noisy labeled instances across all clients in different federated rounds for various detection mechanisms. We observe that energy scores provide an effective mechanism for detection of noisy labeled instances, following a steep curve and outperforming Confidence Learning across all considered noise profiles. In particular, after $30$ federated rounds, energy score is within a close proximity of it's maximum AUC score across all federated rounds. Therefore, for the results reported in the remaining of Section~\ref{ssec:results}, we use a fixed $R_w$=$30$ for all model-based approaches to identification of noisy clients.

\setlength{\textfloatsep}{0.1cm}
\begin{table}[b]
    \centering \tiny
    \caption{\footnotesize{AUC scores on detection of noisy labeled data in CIFAR-10 and SpeechCommands. Considered approaches include Confidence Learning (CL)~\citep{cl}, FedCorr~\citep{4}, while Softmax and Energy~\cite{energy_paper} correspond to our model confidence approach, when the corresponding scoring function is used. Embeddings account for the proposed embedding-based discovery of noisy labels during the initialization phase of FL. Federated parameters are set to $M$=30, $F$=80\%, $\sigma$=25\%,
    while $R$, $E$ and $q$ are fixed to 150,5,1 and 30,1,80 for FedCorr and all other approaches, respectively.}\label{tab:noise_esti_auc}}
    \begin{minipage}[t]{.535\textwidth}
        \centering             
        \begin{tabular}{llccccc}
            \toprule
            \multicolumn{2}{c}{\textbf{\begin{tabular}[c]{@{}c@{}}Noise Profile\\($n_l\%$ , $n_s\%$)\end{tabular}}} & \multicolumn{1}{c}{\textbf{CL}} & \multicolumn{1}{c}{\textbf{FedCorr}} & \multicolumn{1}{c}{\textbf{Softmax}} & \multicolumn{1}{c}{\textbf{Energy}} & \multicolumn{1}{c}{\textbf{Embeddings}} \\ \midrule
            \multirow{2}{*}{$n_l$=$40$}  & $n_s$=$0$  & 60.37 & 67.32 & 57.25 & 68.80 & \textbf{93.55} \\
                                         & $n_s$=$40$ & 58.13 & 64.51 & 55.27 & 65.07 & \textbf{93.43} \\
            \multirow{2}{*}{$n_l$=$70$}  & $n_s$=$0$  & 52.23 & 59.11 & 52.81 & 57.71 & \textbf{91.63} \\
                                         & $n_s$=$70$ & 53.73 & 54.08 & 51.89 & 54.40 & \textbf{62.29} \\ \bottomrule
        \end{tabular}
        \vskip.8\baselineskip
        (a) CIFAR-10 
    \end{minipage}%
    \hfill
    \begin{minipage}[t]{.455\textwidth}
        \centering
        \begin{tabular}{llcccc}
            \toprule
            \multicolumn{2}{c}{\textbf{\begin{tabular}[c]{@{}c@{}}Noise Profile\\($n_l\%$ , $n_s\%$)\end{tabular}}} & \multicolumn{1}{c}{\textbf{CL}} & \multicolumn{1}{c}{\textbf{Softmax}} & \multicolumn{1}{c}{\textbf{Energy}} & \multicolumn{1}{c}{\textbf{Embeddings}} \\ \midrule
            \multirow{2}{*}{$n_l$=$40$}     & $n_s$=$0$  & 62.37 & 59.12 & 70.21 & \textbf{83.16} \\
                                            & $n_s$=$40$ & 61.85 & 57.98 & 69.59 & \textbf{83.07} \\
            \multirow{2}{*}{$n_l$=$70$}     & $n_s$=$0$  & 54.37 & 52.93 & 63.01 & \textbf{82.57} \\
                                            & $n_s$=$70$ & 54.22 & 52.02 & 59.33 & \textbf{60.11} \\ \bottomrule
        \end{tabular}
        \vskip.8\baselineskip 
        (b) SpeechCommands 
    \end{minipage}
\end{table}
\setlength{\floatsep}{0.1cm}

\begin{figure}[t]
    \centering
    \subfigure[Uniform label noise distribution with $n_l$=40\%.]{\label{fig:detb_a} \includegraphics[width=.44\textwidth]{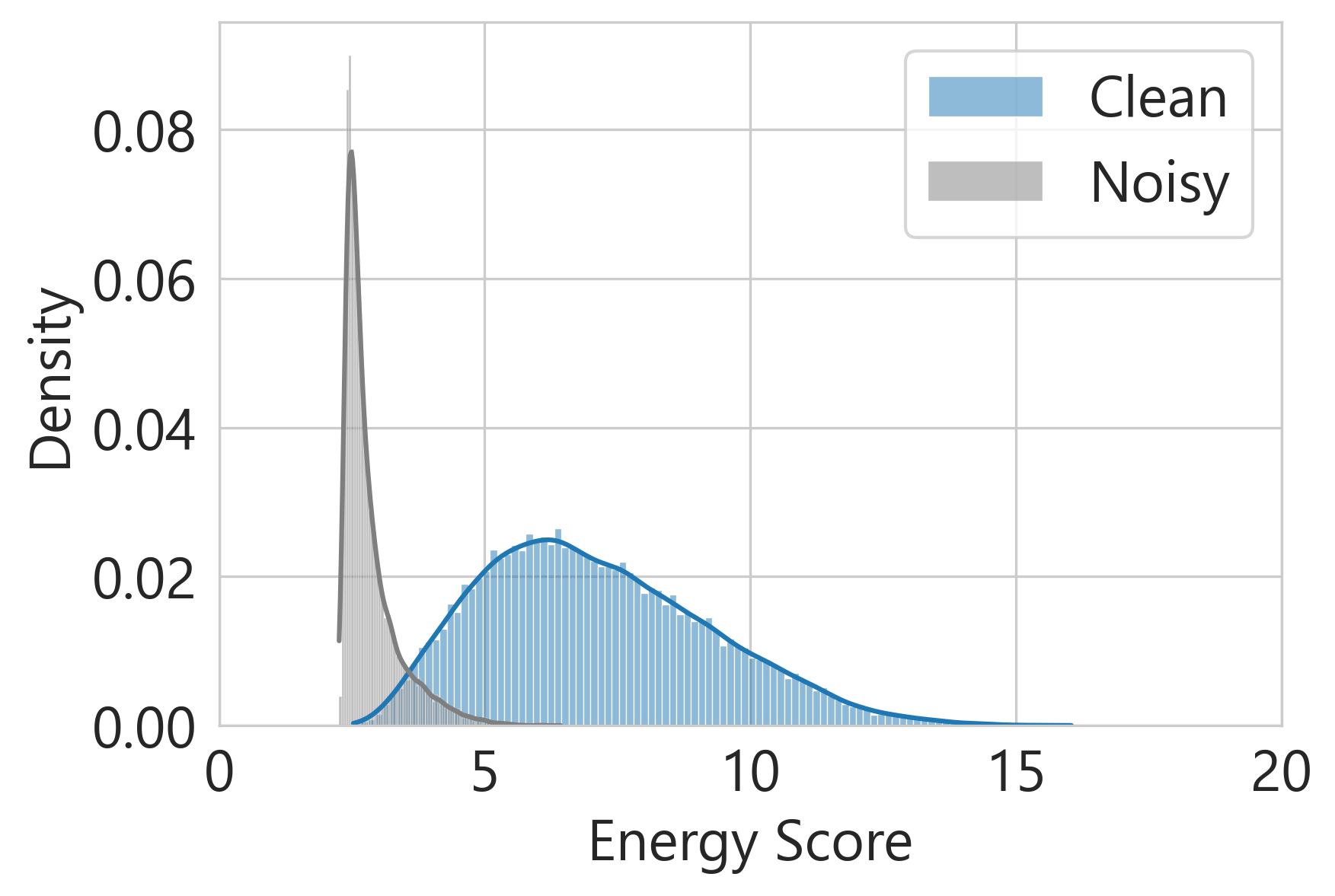}}\hfill%
    \subfigure[Random label noise distribution with $n_l$=40\%.]{\label{fig:detb_b} \includegraphics[width=.44\textwidth]{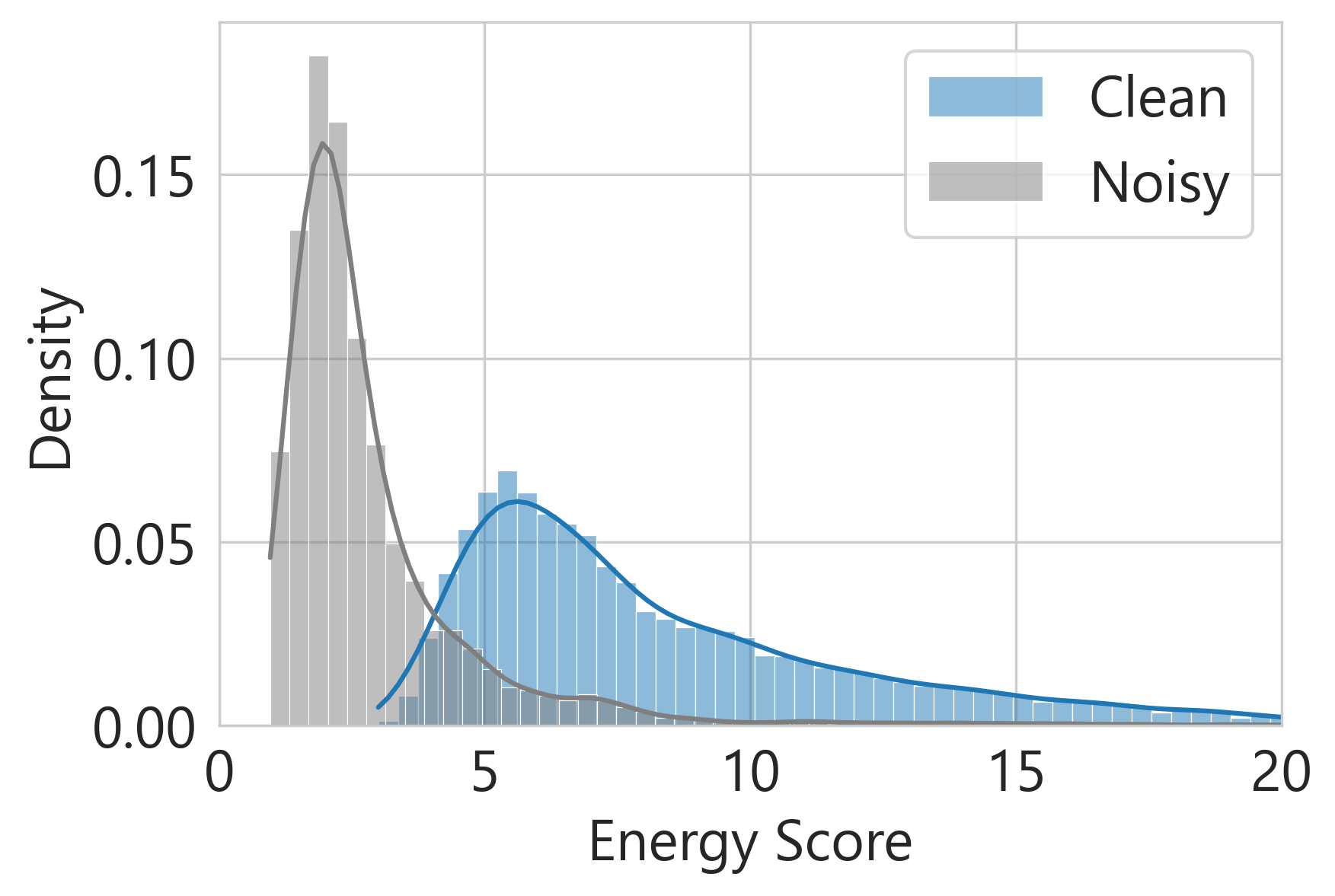}}\hfill%
    \vspace*{-0.4cm}
    \caption{\footnotesize{Energy scores computed for CIFAR-10 on $R$=$30$ using client's local models with $F$=80\%. ``\textit{Clean}'' and ``\textit{noisy}'' terms refers to the presence of label noise in each client's dataset. In (a) a uniform noise with $n_l$=40\% is considered across all ``\textit{noisy}'' clients, and in (b) label noise is follows random distribution across clients, while both (a) and (b) maintain the same number of noisy labeled instances. A clear separation between the scores computed from clean and noisy clients is evident. Note that labels are not explicitly used to compute these scores.}\label{fig:detb}}
\end{figure}

In Table~\ref{tab:noise_esti_auc}, the AUC scores of all considered approaches is reported for CIFAR-10 and SpeechCommands. Comparing all model-based approaches with embedding-based one, we note that the latter's performance is superior. This is due to the difference in the quality of the noise-free ``\textit{source}'' that is utilized to detect noisy labeled instances on the proposed methods. While embeddings are extracted from a pre-trained model that is completely unrelated to clients' own models or their label noise, the remaining approaches rely on the ``\textit{memorization effect}'' of deep learning models during the early stages to detect noisy instances. Consequently, as the label noise increases in the federated setting, a certain degree of noise will be implicitly introduced during the model aggregation process. This can be observed by the sharp decrease, averaging 7\%, in the AUC score for detecting noisy labeled instances across the two datasets in all model-based approaches when the label noise level ($n_l$) is increased from 40\% to 70\%. However, even in the presence of high levels of label noise, it is still possible to observe significant differences in the generalization capabilities of client models with noisy labels and those with clean labels. This information is valuable for re-weighting the contribution of client models during server-side model aggregation. To illustrate this point, we present histograms of energy scores obtained at $R$=$30$ in Figure~\ref{fig:detb}, showing a clear separation between the scores of ``\textit{clean}'' and ``\textit{noisy}'' clients. Here, we consider two cases of label noise: one with clients sharing the same amount of noise and another with a random label noise distribution, both maintaining the same number of noisy labeled instances.

\setlength{\textfloatsep}{0.1cm}
\setlength{\floatsep}{0.1cm}

\setlength{\textfloatsep}{0.1cm}
\begin{figure}[b]
    \centering
    \subfigure[t-SNE on embeddings for $n_l$=40\% and $n_s$=0\%.]{\label{fig:deta} \includegraphics[width=.44\textwidth]{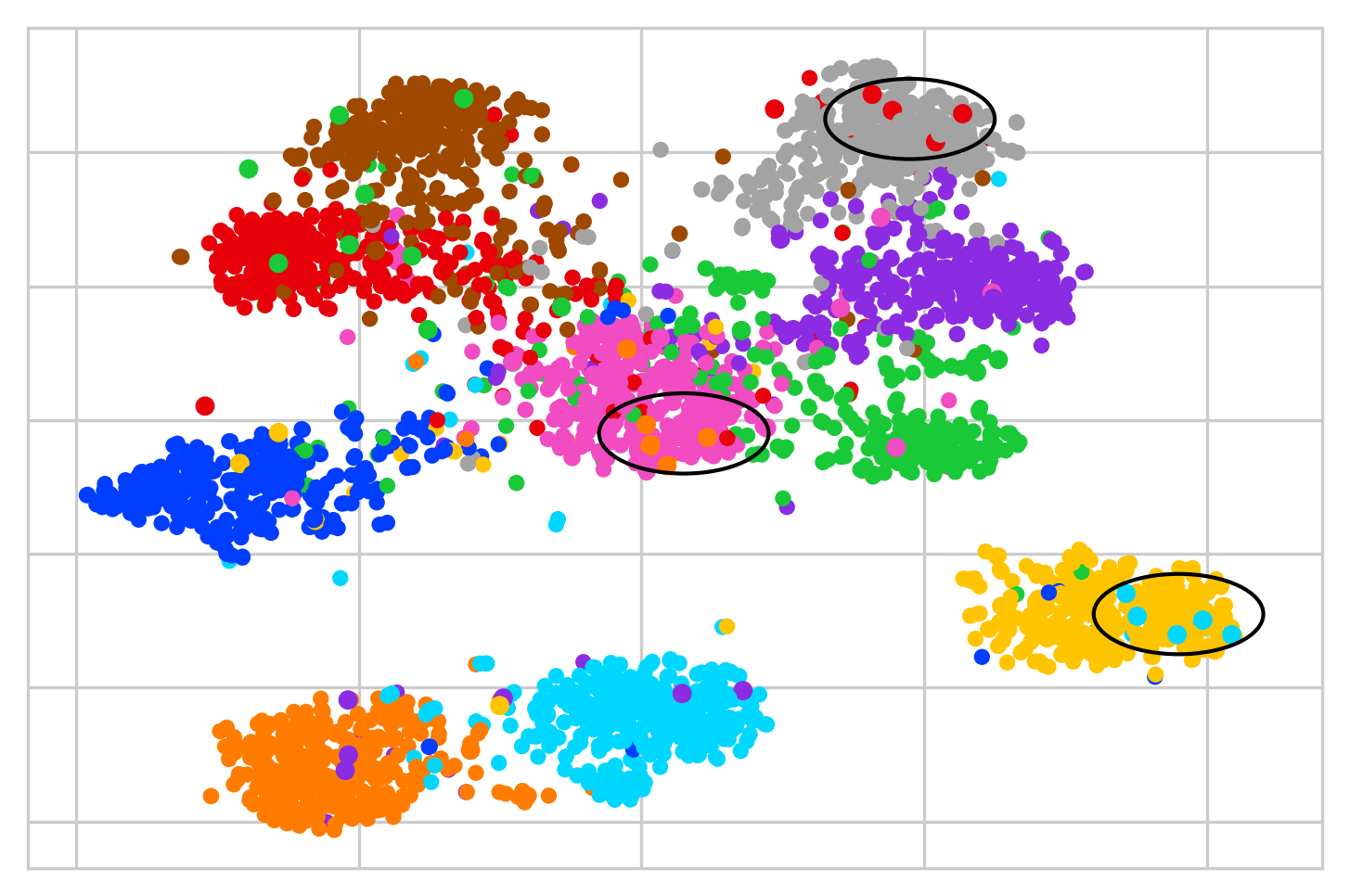}} \hfill%
    \subfigure[t-SNE on embeddings for $n_l$=40\% and $n_s$=0\% after NNC label correction process.]{\label{fig:detaa} \includegraphics[width=.44\textwidth]{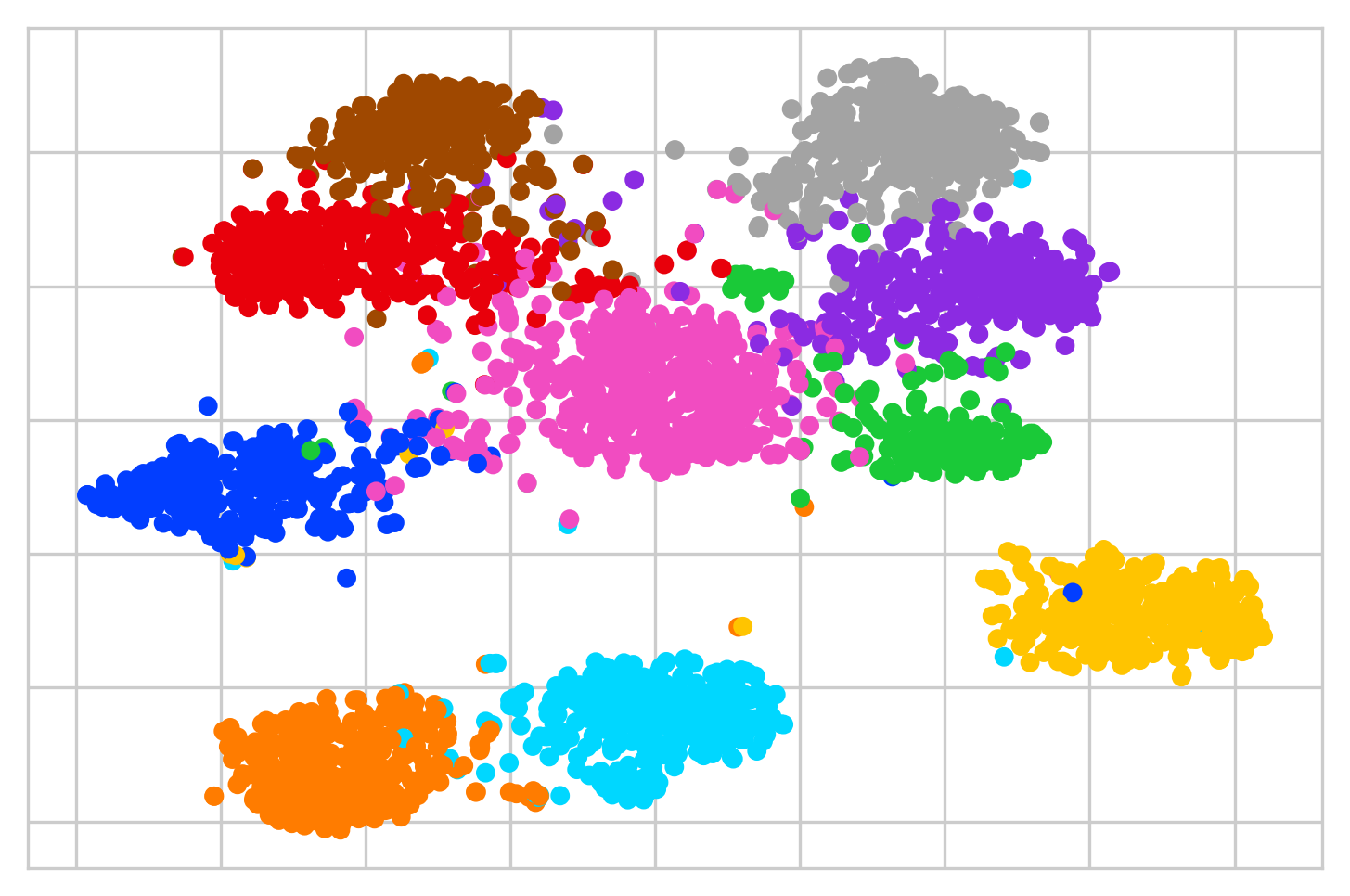}} \hfill%
    \vspace*{-0.4cm}
    \caption{\footnotesize{Intuition behind~\method~ embedding-based approaches for noise level estimation. Figures (a) and (b) show t-SNE on computed embeddings using~\cite{radford2021learning} for CIFAR-10 before and after the label correction using NNC (see Section~\ref{sec:nnc}). Few cases of noisy labeled instances are highlighted with black circles in (a), where examples of one class are incorrectly assigned label from another class. Embedding-based discovery of noisy labels relies on the detection of outlier in a given neighborhood.}\label{fig:noise_estimation_intuition}}
\end{figure}
\setlength{\floatsep}{0.1cm}

On the contrary, the embedding-based discovery approach exhibits greater robustness to the client's noise profile. However, there is an exception in the case of high noise with high sparsity (where $n_l$=$70$\% and $n_s$=$70$/$100$\%). In this scenario, labels belonging to groups of classes become easily confused. Table~\ref{tab:noise_esti_auc} shows that, on average, there is an approximate 30\% decline in the AUC score when $n_l$=$70$\% and $n_s$ increases from $40$\% to $70$\%. While the computed embeddings are unrelated to the noise profile, they amplify the noise level through faulty label estimation. To understand this behavior, we provide an intuition behind how noisy labeled instances are identified using our embedding-based approach in Figure~\ref{fig:noise_estimation_intuition}, where t-SNE visualization is used for CIFAR-10 embeddings extracted using~\citep{radford2021learning}. Noisy labeled instances are indicated by black circles in Figure~\ref{fig:deta}. As discussed in Section~\ref{sec:nnc}, the discovery of noisy labels using the ``\textit{NNC}'' method relies on the detection of outliers in a given neighborhood through majority vote, as shown in Figure~\ref{fig:detaa}. However, in cases of high noise sparsity, labels from a group of classes become mixed, resulting in a concentration of noisy labels from a particular class in a given neighborhood. This amplifies the noise level and leads to incorrect label estimation. For example, in the neighborhood of a cat's image (ground truth label) in the embedding space, where 7 out of the 10 samples are mislabeled as dogs due to high noise sparsity between cats and dogs, ``\textit{NNC}'' computes the estimated label based on the dominant label in the neighborhood, resulting in a dog's label being assigned. To overcome this issue, the ``\textit{AKD}'' method (see Section~\ref{sec:akd}) can be used to exploit embeddings for distillation rather than fixing labels directly, as demonstrated in Section~\ref{sssec:main_experiments}.

\subsubsection{\textbf{Comparison of~\method~against existing techniques}}\label{sssec:main_experiments}\hfill

\setlength{\textfloatsep}{0.1cm}
\begin{table*}[t]
    \scriptsize \centering
    \caption{\footnotesize{Performance evaluation of~\method~on different datasets against a range of baselines. Average accuracy over three distinct trials on test set is reported. Supervised refers to standard FedAvg~\citep{fedavg} training process, while LS and Bi-Temp denote the use of Label Smoothing~\citep{ls} regularization and Bi-Tempered loss~\citep{bTL}, respectively. We include FedCorr~\citep{4} on all vision-based datasets, while CL corresponds to the Confidence Learning~\citep{cl} technique. In AKD, we report the performance for the case of embeddings as source of supervision. The accuracies for AKD with globally aggregated model’s outputs (i.e., logits) as supervisory signals are provided in Table~\ref{tab:main_res_extra} of the Appendix. Federated parameters are set to $R$=200, $M$=30, $F$=80\%, $E$=1, $q$=80\%, and $\sigma$=25\%.}} \label{tab:main_res}
     \vspace*{-0.2cm}

        \begin{tabular}{lllccccccccc}
            \toprule
            \multicolumn{3}{l}{\textbf{Noise ($n_l$)}}                                & 0.0 & \multicolumn{4}{c}{0.4} & \multicolumn{4}{c}{0.7} \\ \cmidrule[0.3pt](lr{.75em}){5-8} \cmidrule[0.3pt](lr{.75em}){9-12}
            \multicolumn{3}{l}{\textbf{Sparsity ($n_s$)}}                                 & 0.0 & 0.0 & 0.4 & 0.7 & 1.0 & 0.0 & 0.4 & 0.7 & 1.0 \\ \midrule
            \multirow{8}{*}{\textbf{CIFAR-10}}   
                                        & \multicolumn{2}{l}{Centralized}               & 91.52 & 76.61 & 76.98 & 76.91 & 69.47 & 58.83 & 58.33 & 57.42 & 45.06 \\ 
                                        \cmidrule[0.1pt](lr{.75em}){2-12}
                                        & \multirow{4}{*}{FedAvg}       & Supervised    & 78.52 & 68.77 & 67.05 & 67.31 & 67.92 & 57.65 & 56.94 & 56.81 & 63.54 \\
                                        &                               & LS            & 73.91 & 68.63 & 64.91 & 64.02 & 64.68 & 58.08 & 56.53 & 56.21 & 62.04 \\
                                        &                               & Bi-Temp.      & 75.29 & 66.18 & 65.66 & 66.58 & 67.71 & 56.75 & 57.61 & 58.39 & 63.74 \\
                                        &                               & CL            & 73.84 & 68.96 & 67.65 & 68.98 & 68.03 & 59.25 & 60.28 & 61.21 & 64.99 \\
                                        & \multirow{1}{*}{FedCorr}      &               & 76.46 & 70.79 & 71.14 & 70.49 & 71.22 & 67.27 & 68.17 & 67.76 & 65.23 \\
                                        \cmidrule[0.1pt](lr{.75em}){2-12}
                                        & \multirow{3}{*}{\method}      & NNC           & 75.29 & \textbf{73.68} & \textbf{73.64} & \textbf{74.23} & \textbf{71.44} & \textbf{74.76} & \textbf{72.63} & 64.49 & 54.59 \\
                                        &                               & AKD           & \textbf{83.55} & 69.72 & 69.18 & 70.26 & 70.55 & 68.43 & 68.06 & 69.74 & \textbf{68.39} \\
                                        &                               & NA-FedAvg     & 76.41 & 69.52 & 70.73 & 70.61 & 71.01 & 65.34 & 66.04 & \textbf{68.11} & 67.92 \\ 
                                        \midrule

            \multirow{8}{*}{\textbf{\begin{tabular}[c]{l}Fashion\\MNIST\end{tabular}}}    
                                        & \multicolumn{2}{l}{Centralized}                  & 91.85 & 83.56 & 86.76 & 86.18 & 78.26 & 63.87 & 60.96 & 61.95 & 40.32 \\
                                        \cmidrule[0.1pt](lr{.75em}){2-12}
                                        & \multirow{4}{*}{FedAvg}    & Supervised          & 86.43 & 82.05 & 83.24 & 83.35 & 81.07 & 58.55 & 56.06 & 57.11 & 59.37 \\
                                        &                               & LS               & 84.86 & 82.08 & 82.07 & 81.88 & 79.84 & 59.38 & 56.24 & 56.95 & 55.66 \\
                                        &                               & Bi-Temp.         & 84.61 & 81.93 & 81.15 & 81.84 & 81.46 & 57.93 & 57.35 & 58.31 & 59.24 \\
                                        &                               & CL               & 83.91 & 82.82 & 83.29 & 83.76 & 81.91 & 59.48 & 57.97 & 57.33 & 60.89 \\
                                        & \multirow{1}{*}{FedCorr}      &                  & 86.51 & 84.93 & 84.37 & 83.71 & 82.55 & 79.47 & 79.28 & 75.62 & 74.01 \\
                                        \cmidrule[0.1pt](lr{.75em}){2-12}
                                        & \multirow{3}{*}{\method}      & NNC              & 80.29 & \textbf{86.81} & \textbf{87.22} & \textbf{87.77} & \textbf{83.13} & \textbf{85.38} & \textbf{84.91} & \textbf{76.88} & 44.48 \\
                                        &                               & AKD              & \textbf{86.91} & 83.97 & 84.06 & 83.53 & 82.35 & 80.77 & 78.63 & 80.41 & 78.64 \\
                                        &                               & NA-FedAvg        & 86.39 & 83.59 & 84.28 & 84.45 & 83.02 & 78.22 & 78.68 & 76.17 & \textbf{79.46} \\ 
                                        \midrule

            \multirow{8}{*}{\textbf{\begin{tabular}[c]{l}Path\\MNIST\end{tabular}}}  
                                        & \multicolumn{2}{l}{Centralized}                  & 90.65 & 81.16 & 80.92 & 81.02 & 78.05 & 58.33 & 59.82 & 57.75 & 47.89 
                                        \\ \cmidrule[0.1pt](lr{.75em}){2-12}
                                        & \multirow{4}{*}{FedAvg}       & Supervised       & 87.05 & 78.82 & 77.06 & 76.68 & 77.03 & 54.74 & 52.49 & 53.22 & 58.61 \\
                                        &                               & LS               & 84.13 & 79.62 & 76.96 & 74.57 & 74.9 & 56.17 & 52.06 & 52.31 & 53.46  \\
                                        &                               & Bi-Temp.         & 83.09 & 78.03 & 77.23 & 77.61 & 76.67 & 55.89 & 55.74 & 56.15 & 58.54 \\
                                        &                               & CL               & 84.31 & 78.97 & 79.09 & 77.26 & 81.02 & 59.69 & 55.88 & 54.29 & 60.21 \\
                                        & \multirow{1}{*}{FedCorr}      &                  & 86.01 & 81.07 & 81.32 & 80.06 & 78.03 & 77.75 & 76.81 & 74.53 & 71.18 \\
                                        \cmidrule[0.1pt](lr{.75em}){2-12}
                                        & \multirow{3}{*}{\method}      & NNC              & 84.45 & \textbf{82.76} & \textbf{82.97} & \textbf{82.01} & 78.97 & \textbf{80.13} & \textbf{82.74} & 78.78 & 41.53 \\
                                        &                               & AKD              & \textbf{87.82} & 82.42 & 82.83 & 81.46 & \textbf{83.26} & 79.94 & 78.63 & 78.31 & \textbf{76.02} \\
                                        &                               & NA-FedAvg        & 85.96 & 80.52 & 81.06 & 81.36 & 78.35 & 76.69 & 75.85 & \textbf{79.64} & 72.84 \\ 
                                        \midrule

            \multirow{8}{*}{\textbf{\begin{tabular}[c]{l}EuroSAT\end{tabular}}}  
                                        & \multicolumn{2}{l}{Centralized}                   & 95.15 & 87.09 & 86.95 & 86.04 & 79.50 & 71.81 & 71.85 & 70.28 & 54.96 \\ 
                                        \cmidrule[0.1pt](lr{.75em}){2-12}
                                        & \multirow{4}{*}{FedAvg}       & Supervised        & 95.07 & 86.68 & 86.97 & 84.98 & 82.98 & 71.70 & 70.29 & 69.37 & 71.99 \\
                                        &                               & LS                & 89.13 & 87.17 & 86.63 & 82.66 & 82.18 & 72.19 & 69.52 & 68.92 & 68.61  \\
                                        &                               & Bi-Temp.          & 89.94 & 85.33 & 85.92 & 84.32 & 82.32 & 71.77 & 70.92 & 70.03 & 72.24 \\
                                        &                               & CL                & 90.33 & 87.94 & 88.33 & 86.18 & 84.05 & 74.29 & 71.68 & 71.12 & 72.57 \\ 
                                        & \multirow{1}{*}{FedCorr}      &                   & 90.98 & 89.29 & 89.02 & 87.53 & 86.17 & 83.53 & 83.26 & 81.96 & 80.11 \\
                                        \cmidrule[0.1pt](lr{.75em}){2-12}
                                        & \multirow{3}{*}{\method}      & NNC               & 88.23 & \textbf{93.18} & \textbf{93.11} & \textbf{92.77} & 85.55 & \textbf{93.48} & \textbf{92.72} & 69.51 & 34.59 \\
                                        &                               & AKD               & \textbf{94.05} & 90.38 & 90.02 & 91.11 & \textbf{89.33} & 88.64 & 89.06 & \textbf{88.33} & \textbf{87.56} \\
                                        &                               & NA-FedAvg         & 90.77 & 88.49 & 89.17 & 88.56 & 85.86 & 83.34 & 83.68 & 84.46 & 86.91 \\
                                        \midrule

            \multirow{8}{*}{\textbf{\begin{tabular}[c]{l}Speech\\Commands\end{tabular}}}   
                                        & \multicolumn{2}{l}{Centralized}                  & 96.68 & 90.33 & 90.31 & 90.84 & 84.84 & 84.95 & 83.31 & 82.65 & 60.41 
                                        \\ \cmidrule[0.1pt](lr{.75em}){2-12}
                                        & \multirow{4}{*}{FedAvg}       & Supervised       & \textbf{96.31} & 81.83 & 82.53 & 82.44 & 80.33 & 72.34 & 70.34 & 70.89 & 72.39 \\
                                        &                               & LS               & 94.64 & 91.13 & 84.77 & 80.11 & 79.35 & 77.06 & 71.28 & 68.13 & 69.71 \\
                                        &                               & Bi-Temp.         & 96.21 & 82.31 & 81.35 & 82.76 & 82.78 & 73.27 & 71.41 & 72.57 & 70.98 \\
                                        &                               & CL               & 87.12 & 85.45 & 87.97 & 84.34 & 85.54 & 78.34 & 72.92 & 70.07 & 72.81 
                                        \\ \cmidrule[0.1pt](lr{.75em}){2-12}
                                        & \multirow{3}{*}{\method}        & NNC            & 95.79 & \textbf{95.91} & \textbf{95.95} & \textbf{95.97} & \textbf{96.24} & \textbf{96.09} & \textbf{96.11} & \textbf{96.13} & 46.07 \\
                                        &                               & AKD              & 84.82 & 86.42 & 84.83 & 85.46 & 83.26 & 79.94 & 76.63 & 78.31 & 76.02 \\
                                        &                               & NA-FedAvg        & 96.07 & 89.49 & 90.35 & 92.72 & 94.09 & 79.12 & 81.91 & 82.33 & \textbf{80.37} \\
            \bottomrule
    \end{tabular}
\end{table*}
\setlength{\floatsep}{0.1cm}

Here, we compare~\method~to determine the achieved improvements versus other considered baselines, and assess~\method~effectiveness on a wide range of tasks from both vision and audio domains. To this end, we perform experiments on all datasets for a diverse number of noisy profiles, varying both noise level ($n_l$) and sparsity ($n_s$). For a fair comparison, we utilize identical data partitioning and noisy injection schemes in all related experiments. Table~\ref{tab:main_res} provides the accuracy scores on test sets averaged across three independent runs to be robust against differences in randomness involve in training deep neural networks. In the experiments we conducted in centralized setting, models are trained until convergence to obtain the resulting accuracy on a test set, which is presented in the centralized rows of Table~\ref{tab:main_res}. Additionally, for ease of comparison and benchmarking, we include the obtained accuracy for all considered approaches and datasets when no additional noise in injected in data, thus, acting as an upper bound of models' performance. In the case of \textit{AKD}, we report the performance for the case of embeddings as source of supervision in Table~\ref{tab:main_res}, while the accuracies for identical experiments performed for \textit{AKD} with globally aggregated model's outputs (i.e., logits) as supervisory signal are provided in Table~\ref{tab:main_res_extra} of the Appendix.

In Table~\ref{tab:main_res}, we observe our approaches can improve the model's performance compared to standard \textit{FedAvg} across all datasets significantly. Consequently, we can conclude that~\method~can be applied in a federated environment with noisy labels to boost the performance, independent of the learning task or inputs' modality. In particular, comparing the rows for $n_l$=$70$\% we note an increase of $22.67$\% in accuracy on average using~\method~across the considered tasks compared to the standard federated model. 

Observing the results obtained with the baselines, we notice inconsistencies in the performance of both LS and Bi-Tempered loss across a wide range of noise profiles. LS shows effectiveness for low sparsity label noises, while Bi-Tempered loss produces desirable results only for large levels of label noise. CL, on the other hand, demonstrates more stable performance across diverse label noise settings; yet, it's overall improvement does not exceed an average of 4\% across the tasks. In comparison, FedCorr, a label correction approach designed for FL, can effectively handle label noise across most settings, particularly under low sparsity label noise. However, its performance diminishes as the noise sparsity increases, resulting in less accurate pseudo-labels. On the contrary,~\method~performance is stable across a wide range of noise profiles and provide significant improvement in model's generalization capability. From our proposed approaches, we note that \textit{NNC} provides the largest improvement on model's performance across distinct noise profiles, with the exception of the special cases of ``\textit{class-flipping}'' on high levels of noise ($n_l$=$40$/$70$\% and $n_s$=$100$\%). In such cases, \textit{NNC} is unable to use the computed embeddings to perform the label correction process, as discussed in Section~\ref{sssec:noisy_samples}, while \textit{AKD} can adequately handle these cases to properly utilize the embeddings and overcome the effect of label noise. Apart from the particular ``\textit{class-flipping}'' noise, \textit{AKD}'s embedding-based distillation approach could possibly surpass the performance of \textit{NNC}, in case FL runs for a longer period ($R$>$200$) at the expense of increased computational overhead. This is evident when no noise is present, where \textit{AKD} surpass all other approaches in federated setting for all datasets from the vision domain, showcasing that it is a ``\textit{cleaner}'' supervision signal with no correlation to existing labels. Furthermore, we observe that \textit{NA-FedAvg} can remain effective across all considered noise profiles, while introducing minimal computational overhead and client-side modifications during the FL training process. In particular, \textit{NA-FedAvg} performance is in pair with FedCorr, where the latter requires extensive federated rounds and a computationally expensive process involving training of GMM to produce a label noise estimate per-client. Therefore, \textit{NA-FedAvg} can be an ideal alternative for clients with minimal computational and storage resources to train federated models under label noise.

\subsubsection{\textbf{Evaluation of~\method~ with different model architectures.}}\label{sssec:diverse_model}\hfill

Our proposed methods for dealing with label noise in FL attain high performance across a wide range of classification tasks. While in \textit{NNC} the label correction process does not utilize clients' model's, both \textit{AKD} and \textit{NA-FedAvg} approaches involves clients' models. To ensure that~\method~efficacy is not related to a specific network architecture or model optimization, we perform experiments on CIFAR-10 and EuroSAT datasets, where, we replace the ResNet-20 model with a recent convolutional-based neural network, named ConvMixer~\citep{convmixer}. In particular, we choose ConvMixer-128/4, with a kernel size of $2$, a patch size of $5$ and approximately $0.1$M parameters. Such small memory footprint and low complexity, render ConvMixer-128/4 ideal for on-device learning. In our experimentation with ConvMixer, we use Adam optimizer with learning rate of $0.001$, similar to the original paper~\citep{convmixer}.

\setlength{\textfloatsep}{0.1cm}
\begin{table}[t]
    \scriptsize \tiny
    \caption{\footnotesize{Performance evaluation of~\method~with ConvMixer-128/4~\citep{convmixer} on CIFAR-10 and EuroSAT. Average accuracy over three distinct trials on test set is reported. In AKD, we report the performance for the case of embeddings as source of supervision. Federated parameters are set to $R$=200, $M$=30, $F$=80\%, $E$=1, $q$=80\%, and $\sigma$=25\%.}\label{tab:conv_mixer}}
    \vspace*{-0.2cm}
    \begin{minipage}[t]{0.495\hsize} \centering \resizebox{1.0\hsize}{!}{%
        \begin{tabular}{lccccccccc}
            \toprule
            \textbf{Noise ($n_l$)}      & 0.0 & \multicolumn{4}{c}{0.4} & \multicolumn{4}{c}{0.7} \\ \cmidrule[0.3pt](lr{.75em}){3-6} \cmidrule[0.3pt](lr{.75em}){7-10}
            \textbf{Sparsity ($n_s$)}   & 0.0 & 0.0 & 0.4 & 0.7 & 1.0 & 0.0 & 0.4 & 0.7 & 1.0 \\ 
            \midrule
            Centralized           & 91.26 & 76.61 & 76.98 & 76.31 & 69.47 & 68.83 & 68.33 & 67.42 & 55.06 \\ \cmidrule[0.1pt](lr{.75em}){1-10}
            FedAvg~\citep{fedavg} & 82.04 & 74.64 & 73.26 & 74.28 & 74.41 & 63.46 & 63.94 & 64.36 & 67.25 \\ \cmidrule[0.1pt](lr{.75em}){1-10}
            NNC                   & 77.73 & \textbf{76.53} & \textbf{77.44} & \textbf{77.82} & \textbf{76.33} & \textbf{76.82} & \textbf{76.59} & 70.46 & 30.14 \\
            AKD                   & \textbf{82.25} & 74.31 & 74.18 & 75.04 & 75.18 & 71.89 & 71.63 & \textbf{72.64} & 70.09 \\
            NA-FedAvg             & 81.90 & 75.85 & 74.79 & 76.92 & 75.74 & 68.61 & 68.97 & 67.66 & \textbf{70.18} \\ 
            \bottomrule
        \end{tabular}%
        }
        \vskip.8\baselineskip
        (a) CIFAR-10 
    \end{minipage}%
    \hfill
    \begin{minipage}[t]{0.495\hsize} \centering \resizebox{1.0\hsize}{!}{%
        \begin{tabular}{lccccccccc}
            \toprule
            \textbf{Noise ($n_l$)}          & 0.0 & \multicolumn{4}{c}{0.4} & \multicolumn{4}{c}{0.7} \\ \cmidrule[0.3pt](lr{.75em}){3-6} \cmidrule[0.3pt](lr{.75em}){7-10}
            \textbf{Sparsity ($n_s$)}   	& 0.0 & 0.0 & 0.4 & 0.7 & 1.0 & 0.0 & 0.4 & 0.7 & 1.0 \\ 
            \midrule
            Centralized           & 95.01 & 86.93 & 86.74 & 85.32 & 79.34 & 71.43 & 71.25 & 70.17 & 52.91 \\ \cmidrule[0.1pt](lr{.75em}){1-10}
            FedAvg~\citep{fedavg} & 92.31 & 82.42 & 81.97 & 82.27 & 80.14 & 68.32 & 67.91 & 68.03 & 69.09 \\ \cmidrule[0.1pt](lr{.75em}){1-10}
            NNC                   & 90.11 & \textbf{90.76} & \textbf{90.34} & \textbf{90.49} & 85.01 & \textbf{89.56} & \textbf{89.27} & 78.16 & 31.51 \\
            AKD                   & \textbf{93.52} & 90.07 & 89.27 & 89.84 & \textbf{88.07} & 84.28 & 84.46 & \textbf{83.82} & \textbf{83.51} \\
            NA-FedAvg             & 89.39 & 86.93 & 86.72 & 87.16 & 86.82 & 80.74 & 79.26 & 79.65 & 80.31 \\
            \bottomrule
        \end{tabular}%
        }
        \vskip.8\baselineskip 
        (b) EuroSAT 
    \end{minipage}
\end{table}
\setlength{\floatsep}{0.1cm}

From the results presented in Table~\ref{tab:conv_mixer}, we note that~\method~retains high efficacy, even if a different architecture is used. In particular, for $n_l$=$70$\%, we notice an increase of {\color{mark} $5.32$\%} in accuracy on average using~\method~compared to the standard ``\textit{FedAvg}'', with all three of proposed approaches following similar performance gains to the ones observed in Table~\ref{tab:main_res}. This indicates that both of our embeddings and model confidence based methods can be employed in FL to mitigate the effect of label noise and improve model's generalizability, irrespective of the architecture of federated model to be learned.

\subsubsection{\textbf{Effectiveness of~\method~ across diverse label noise settings}}\label{sssec:diverse_noise}\hfill

So far, in our evaluation, we considered diverse noise settings, assuming that``\textit{clean}'' clients, holding well-annotated data, are present during the FL procedure. In this subsection, we assess the efficacy of~\method, while we relax our assumption regarding the presence of ``\textit{clean}'' clients. To this end, we conduct further experiments on CIFAR-10, where the percentage of noisy clients, and the amount of well-annotated data present on ``\textit{clean}'' clients are varied. 

\hfill\\ \textbf{Varying number of noisy clients:} With the client's labeling systems and user's willingness (or ability) to perform a correct annotation process significantly varies in a federated environment, the number of noisy (and clean) clients present in the FL process can fluctuate drastically. In this ablation study, we conduct experiments to determine the effect of the number of noisy clients on ~\method~performance. To this end, we perform experiments by varying the percentage of noisy clients ($F$) from $25$\% up to $100$\% for $n_l$=$40$\% utilizing both \textit{NNC} and \textit{NA-FedAvg}. In this way, we are able to assess the performance of our methods for estimating label noise (namely, using embeddings and model's predictions confidence). It is important to note that $F$=$100$\% corresponds to a scenario, where all clients contain $n_l$\% of label noise in their locally stored data (not to be confused with a completely noisy dataset). 

\setlength{\textfloatsep}{0.1cm}
\begin{table}[t]
    \scriptsize \centering
    \caption{\footnotesize{Performance evaluation of~\method~when varying the percentage of noisy clients ($F$) for $n_l$=40\% on CIFAR-10. Average accuracy over three distinct trials on test set is reported. Federated parameters are set to $R$=200, $M$=30, $E$=1, $q$=80\%, and $\sigma$=25\%.} \label{tab:vary_noisy_clients}}
    \vspace*{-0.2cm}
    \resizebox{0.75\hsize}{!}{%
        \begin{tabular}{ccccccccc}
            \toprule
            \multirow{2}{*}{\textbf{\% Noisy Clients ($F$)}} & \multicolumn{4}{c}{\textbf{\textit{NNC}}} & \multicolumn{4}{c}{\textbf{\textit{NA-FedAvg}}} \\ \cmidrule[0.3pt](lr{.75em}){2-5} \cmidrule[0.3pt](lr{.75em}){6-9}
                                                                & $n_s$=$0\%$  & $n_s$=$40\%$  & $n_s$=$70\%$  & $n_s$=$100\%$  & $n_s$=$0\%$  & $n_s$=$40\%$  & $n_s$=$70\%$ & $n_s$=$100\%$ \\ \midrule            
            \textbf{25}     & 73.80 & 74.84 & 74.18 & 72.84   & \textbf{75.99} & \textbf{75.97} & \textbf{76.59} & \textbf{76.02} \\
            \textbf{50}     & \textbf{74.54} & \textbf{73.59} & \textbf{74.97} & 74.23   & 73.27 & 73.45 & 72.99 & \textbf{74.36} \\
            \textbf{75}     & \textbf{73.50} & \textbf{74.49} & \textbf{74.96} & \textbf{73.95}   & 70.99 & 70.74 & 71.48 & 70.80 \\
            \textbf{100}    & \textbf{73.76} & \textbf{73.41} & \textbf{73.14} & \textbf{72.49}   & 64.02 & 62.62 & 62.72 & 64.22 \\
            \bottomrule
    \end{tabular}%
    }
\end{table}
\setlength{\floatsep}{0.1cm}

We present our findings in Table~\ref{tab:vary_noisy_clients}, where we note that the number of noisy clients has a relatively low impact on the ability of \textit{NNC} to perform the label correction process, as the embeddings remain unaffected by the noise introduced by those clients. On the contrary, when trained with \textit{NA-FedAvg}, the model's performance deteriorates once all clients become noisy ($F$=$100$\%). However, even with as few as $20$\% of clients being ``\textit{clean}'', the model's performance reaches approximately $70$\% across all noise profiles, as shown in Table~\ref{tab:main_res}. Furthermore, in cases where the label noise is sparse across clients ($F$=$25$\%), we observe that \textit{NA-FedAvg} performance is superior to \textit{NNC}. Although the overall number of noisy labeled instances has a negligible effect on the FL process, a small amount of label noise is introduced to ``\textit{clean}'' clients due to slight imperfections in label estimation via \textit{NNC}.

\hfill\\ \textbf{Varying label noise distribution across devices:} Apart from varying the number of noisy devices in federated setting, the quality of labels present on ``\textit{clean}'' clients can also fluctuate in most pragmatic federated setting. This can happen due to imperfect labeling processes, which frequently occur in real-life, either due to human error or unforeseen mistake in a automated labeling system~\citep{weak_labels}. Therefore, in this ablation study we aim to assess~\method~ performance, where, in addition to noisy clients, ``\textit{clean}'' clients also hold a small amount of label noise. For this purpose, we perform experiments with both NCC and NA-FedAvg for $n_l$=$40$\%, where we introduce a percentage of label noise to ``\textit{clean}'' clients. Note that this percentage is considered a percentage of label noise $n_l$; thus a value of $100$\% corresponds to $n_l$\% of label noise injected in data of ``\textit{clean}'' clients. 

\setlength{\textfloatsep}{0.1cm}
\begin{table}[b]
    \scriptsize \centering
    \caption{\footnotesize{Performance evaluation of~\method~when noise is introduced on ``\textit{clean}'' clients for $n_l$=40\% in CIFAR-10. Average accuracy over three distinct trials on test set is reported. Federated parameters are set to $R$=200, $M$=30, $F$=80\%, $E$=1, $q$=80\%, and $\sigma$=25\%.}} \label{tab:vary_noise_on_clean}
    \vspace*{-0.2cm}
    \resizebox{0.75\hsize}{!}{%
        \begin{tabular}{ccccccccc}
            \toprule
            \multirow{2}{*}{\textbf{\begin{tabular}[c]{c}\% of $n_l$\\on clean clients\end{tabular}}} & \multicolumn{4}{c}{\textbf{\textit{NNC}}} & \multicolumn{4}{c}{\textbf{\textit{NA-FedAvg}}} \\ \cmidrule[0.3pt](lr{.75em}){2-5} \cmidrule[0.3pt](lr{.75em}){6-9} 
                            & $n_s$=$0\%$  & $n_s$=$40\%$  & $n_s$=$70\%$  & $n_s$=$100\%$  & $n_s$=$0\%$  & $n_s$=$40\%$  & $n_s$=$70\%$ & $n_s$=$100\%$ \\ \midrule            
            \textbf{0}      & \textbf{73.68} & \textbf{73.64} & \textbf{74.23} & \textbf{71.44} & 69.52 & 70.73 & 70.61 & 71.01 \\ 
            \textbf{5}      & \textbf{73.53} & \textbf{74.02} & \textbf{75.45} & \textbf{72.24} & 69.17 & 70.46 & 70.02 & 70.87 \\
            \textbf{10}     & \textbf{74.39} & \textbf{74.57} & \textbf{74.70} & \textbf{72.29} & 68.12 & 68.13 & 67.18 & 67.53 \\
            \textbf{25}     & \textbf{73.49} & \textbf{72.94} & \textbf{73.68} & \textbf{72.82} & 67.01 & 66.17 & 64.11 & 66.11 \\
            \textbf{50}     & \textbf{74.25} & \textbf{73.68} & \textbf{72.57} & \textbf{72.98} & 64.88 & 64.02 & 63.32 & 64.57 \\
            \textbf{75}     & \textbf{74.72} & \textbf{74.08} & \textbf{73.48} & \textbf{72.45} & 63.79 & 63.97 & 63.57 & 64.14 \\
            \textbf{100}    & \textbf{74.01} & \textbf{73.08} & \textbf{74.14} & \textbf{73.87} & 64.02 & 62.62 & 62.72 & 64.22 \\ 
            \bottomrule
    \end{tabular}%
    }
\end{table}
\setlength{\floatsep}{0.1cm}

From the results provided in Table~\ref{tab:vary_noise_on_clean}, it can be observed that the performance of NA-FedAvg starts to deteriorate when label noise is injected into the data of ``\textit{clean}'' clients. Specifically, with a noise addition of $0.1 \times n_l$, an average drop in accuracy of 3\% is noted across the noise profiles. This drop occurs as \textit{NA-FedAvg} utilizes ``\textit{clean}'' clients models ability to generalize faster to act as a ``\textit{source}'' for detection of noisy labeled instances. This drop occurs as \textit{NA-FedAvg} relies on the ability of any available ``\textit{clean}'' client's models to generalize faster, acting as a ``\textit{source}'' for the detection of noisy labeled instances. While the overall label noise present in the FL process is similar to that of the experiments reported in Table~\ref{tab:vary_noisy_clients}; here, we introduce label noise to all ``\textit{clean}'' clients instead of flipping them to noisy ones. As a result, a clear separation between the energy scores obtained from the two groups of clients, as illustrated in Figure~\ref{fig:detb}, begins to diminish. Consequently, the ability of \textit{NA-FedAvg} to determine a suitable threshold for determining noisy labelled instances sharply deteriorates, leading to poor generalizability of the obtained model. On the contrary, \textit{NNC} remains effective in performing label correction and producing more generalizable models, even when $n_l$ noise is introduced across all clients in FL. Therefore, \textit{NNC} can be a preferable approach in cases where strong assumptions are made about the noise profile, such as the absence of clients with quality labels. %

\subsubsection{\textbf{Real-world human annotation errors}}\label{sssec:real_noise}\hfill \vspace{0.1pt}

Since synthetic noise often mimic clear structures to enable statistical analyses, it may fails to model complex real-world noise patterns or biases, which impose additional challenges as compared to synthetic label noise. In an effort to evaluate~\method~performance in a more realistic label noise setting, we use the re-annotated versions of the CIFAR-10/100 datasets, which contains real-world human annotation errors, namely CIFAR-10/100N~\citep{cifarn}. In this way, we are able to study~\method~performance with label noise in-the-wild. During the labeling process by human annotators, with the help of Amazon Mechanical Turk, a noise level of approximately $40$\% ($n_l$=$40$\%) was observed. This is evident that human labeling efforts inevitably result in considerable amount of label noise being introduced in the data. In the federated setting, this posses major challenges for user's, who are required to provide well-annotated data to enjoy high-quality FL services, and further necessitates the development of approaches to adequately deal with label noise in FL.

\setlength{\textfloatsep}{0.1cm}
\begin{figure}[t]
    \centering 
    \includegraphics[width=.67\textwidth]{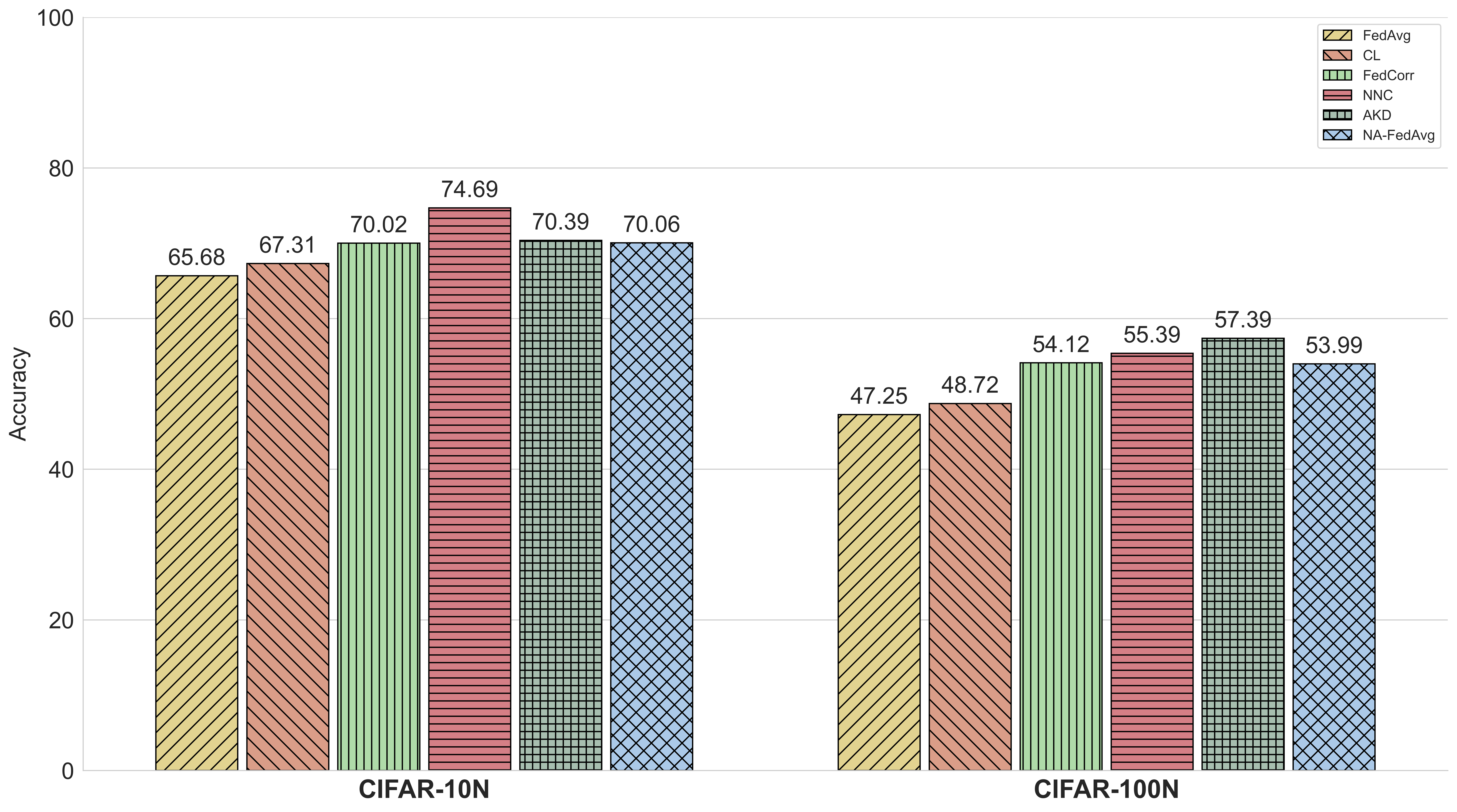}
    \vspace*{-0.2cm}
    \caption{\footnotesize{Evaluation of~\method~in real-world label noise patterns from CIFAR-10/100N. Average accuracy over three distinct trials and on test set is reported for $R$=200/500 for CIFAR-10/100N, respectively. In AKD, we report the performance for the case of embeddings as source of supervision. Remaining federated parameters are set to $M$=30, $E$=1, $q$=80\%, and $\sigma$=25\%.}\label{fig:real_noise}}
\end{figure}
\setlength{\floatsep}{0.1cm}

We perform experiments on both CIFAR-10/100N datasets utilizing~\method~approaches, and include standard FedAvg with same training configuration, to clearly illustrate performance gain of our approach. For a fair comparison, we included CL and FedCorr performance in identical experiments, as these were the two most prominent approaches from Table~\ref{tab:main_res}. While the train set of CIFAR-10/100N was annotated by humans, the original \textit{noise-free} test set of both datasets is used for evaluation. Furthermore, we randomly distribute the data across clients; thus no clear group of ``\textit{clean}'' clients is considered in this case, which make the learning even more challenging. From the CIFAR-10N results presented in Figure~\ref{fig:real_noise}, we note that~\method~retains it's effectiveness, while moving from synthetic to real-world noise patterns. In particular, model's recognition rate remains within $2$\% to the ones reported in Table~\ref{tab:main_res} for $n_l$=$40$\%. In the case of CIFAR100N, where the complexity of the task is increased due to large number of classes, we train all models for $500$ federated rounds ($R$=$500$). In this case, we observe that embedding-based supervision via \textit{AKD} outperforms the remaining approaches, validating that \textit{AKD}'s performance can be especially beneficial if longer federated training is possible. Subsequently, we note that~\method~is able to largely address the challenges introduced from real-life noise patterns and is able to improve the recognition rate by $9$\%, compared to the standard FL process. Therefore, our approaches can be useful for everyday users, who utilize FL services and are requested to provide labels for their data, either explicitly or through their own activities (e.g., the next word prediction in Gboard~\citep{KeywordSpottingIID}), which are inherently noisy.

\section{Conclusions}

We study the pragmatic problem of label noise under the federated setting. In the distributed scenario, clients' well-annotated examples are sparse due to flaws in the labeling process, which originate either from deficient automatic-labeling techniques or users' mistakes. To aggravate this problem, label noise in the federated setting can substantially differ among clients, as it is closely coupled to a number of client-dependent sources, such as the discrepancy between clients’ labeling systems or the difference in clients’ expertise and willingness to label data correctly. Due to this reason, and the scarcity of data in FL, most centralized approaches' (to handle noisy labels) deteriorate in the federated setting, while the limited FL approaches dealing with noisy labels introduce extensive overhead on client-side or rely on server-side clean data availability. To address the lack of computationally efficient ways to deal with label noise during learning on-device models without relying on any additional data, we present~\method~framework, where we propose three different approaches, each operating in a distinct phase of the FL process, providing label noise solutions in FL for diverse device compute characteristics.
Apart from noise mitigation~\method~provide a mechanism to perform label correction. Despite the simplicity of our approaches, namely \textit{NNC}, \textit{AKD} and \textit{NA-FedAvg}, we demonstrate that they can address noisy labeled instances across a wide range of label noise settings. We evaluate~\method~on several publicly available datasets, comparing its performance with several baselines in the federated setting. The models' generalization we achieve is consistently superior to the considered baselines, while an evaluation of~\method~with in-the-wild label noise data, showcase~\method~effectiveness under complex real-world label noise patterns and real-life scenarios. In addition, the minimal communication footprint of\method~in the FL process effectively mitigates additional communication overhead typically associated with large-scale FL applications, making it a viable choice for deployments in various real-world applications. On that note, \textit{NA-FedAvg} will require all clients participation to compute the energy score in a fixed round (though can be preformed in asynchronously fashion); thus, ensuring accurate assessment of clients' noise levels, an assumption also considered in~\citep{4}.

\small{
\begin{acks}
This work is partially performed in the context of the Distributed Artificial Intelligent Systems project supported by the ECSEL Joint Undertaking. Various icons used in the figures are created by Nanda Diga, Soremba, Andriwidodo, Product Pencil, Weltenraser, Kamin, Trevor, Olena, and David from the Noun Project.
\end{acks}
}

\small{
\bibliography{main}


\begin{thebibliography}{52}


\ifx \showCODEN    \undefined \def \showCODEN     #1{\unskip}     \fi
\ifx \showDOI      \undefined \def \showDOI       #1{#1}\fi
\ifx \showISBNx    \undefined \def \showISBNx     #1{\unskip}     \fi
\ifx \showISBNxiii \undefined \def \showISBNxiii  #1{\unskip}     \fi
\ifx \showISSN     \undefined \def \showISSN      #1{\unskip}     \fi
\ifx \showLCCN     \undefined \def \showLCCN      #1{\unskip}     \fi
\ifx \shownote     \undefined \def \shownote      #1{#1}          \fi
\ifx \showarticletitle \undefined \def \showarticletitle #1{#1}   \fi
\ifx \showURL      \undefined \def \showURL       {\relax}        \fi
\providecommand\bibfield[2]{#2}
\providecommand\bibinfo[2]{#2}
\providecommand\natexlab[1]{#1}
\providecommand\showeprint[2][]{arXiv:#2}

\bibitem[Amid et~al\mbox{.}(2019)]%
        {bTL}
\bibfield{author}{\bibinfo{person}{Ehsan Amid}, \bibinfo{person}{Manfred~K.
  Warmuth}, \bibinfo{person}{Rohan Anil}, {and} \bibinfo{person}{Tomer Koren}.}
  \bibinfo{year}{2019}\natexlab{}.
\newblock \showarticletitle{Robust Bi-Tempered Logistic Loss Based on Bregman
  Divergences}.
\newblock  (\bibinfo{year}{2019}).
\newblock
\urldef\tempurl%
\url{https://doi.org/10.48550/ARXIV.1906.03361}
\showDOI{\tempurl}


\bibitem[Arazo et~al\mbox{.}(2019)]%
        {gmm_noise}
\bibfield{author}{\bibinfo{person}{Eric Arazo}, \bibinfo{person}{Diego Ortego},
  \bibinfo{person}{Paul Albert}, \bibinfo{person}{Noel~E. O'Connor}, {and}
  \bibinfo{person}{Kevin McGuinness}.} \bibinfo{year}{2019}\natexlab{}.
\newblock \bibinfo{title}{Unsupervised Label Noise Modeling and Loss
  Correction}.
\newblock
\newblock
\showeprint[arxiv]{1904.11238}~[cs.CV]


\bibitem[Arpit et~al\mbox{.}(2017)]%
        {mem}
\bibfield{author}{\bibinfo{person}{Devansh Arpit},
  \bibinfo{person}{Stanis{\l}aw Jastrz{\k{e}}bski}, \bibinfo{person}{Nicolas
  Ballas}, \bibinfo{person}{David Krueger}, \bibinfo{person}{Emmanuel Bengio},
  \bibinfo{person}{Maxinder~S Kanwal}, \bibinfo{person}{Tegan Maharaj},
  \bibinfo{person}{Asja Fischer}, \bibinfo{person}{Aaron Courville},
  \bibinfo{person}{Yoshua Bengio}, {et~al\mbox{.}}}
  \bibinfo{year}{2017}\natexlab{}.
\newblock \showarticletitle{A closer look at memorization in deep networks}. In
  \bibinfo{booktitle}{\emph{International conference on machine learning}}.
  PMLR, \bibinfo{pages}{233--242}.
\newblock


\bibitem[Bai et~al\mbox{.}(2021)]%
        {es1}
\bibfield{author}{\bibinfo{person}{Yingbin Bai}, \bibinfo{person}{Erkun Yang},
  \bibinfo{person}{Bo Han}, \bibinfo{person}{Yanhua Yang},
  \bibinfo{person}{Jiatong Li}, \bibinfo{person}{Yinian Mao},
  \bibinfo{person}{Gang Niu}, {and} \bibinfo{person}{Tongliang Liu}.}
  \bibinfo{year}{2021}\natexlab{}.
\newblock \showarticletitle{Understanding and Improving Early Stopping for
  Learning with Noisy Labels}. In \bibinfo{booktitle}{\emph{NeurIPS}}.
\newblock


\bibitem[Beutel et~al\mbox{.}(2020)]%
        {Flower}
\bibfield{author}{\bibinfo{person}{Daniel~J Beutel}, \bibinfo{person}{Taner
  Topal}, \bibinfo{person}{Akhil Mathur}, \bibinfo{person}{Xinchi Qiu},
  \bibinfo{person}{Titouan Parcollet}, {and} \bibinfo{person}{Nicholas~D
  Lane}.} \bibinfo{year}{2020}\natexlab{}.
\newblock \showarticletitle{Flower: A Friendly Federated Learning Research
  Framework}.
\newblock \bibinfo{journal}{\emph{arXiv preprint arXiv:2007.14390}}
  (\bibinfo{year}{2020}).
\newblock


\bibitem[Chen et~al\mbox{.}(2020)]%
        {8}
\bibfield{author}{\bibinfo{person}{Yiqiang Chen}, \bibinfo{person}{Xiaodong
  Yang}, \bibinfo{person}{Xin Qin}, \bibinfo{person}{Han Yu},
  \bibinfo{person}{Biao Chen}, {and} \bibinfo{person}{Zhiqi Shen}.}
  \bibinfo{year}{2020}\natexlab{}.
\newblock \bibinfo{title}{FOCUS: Dealing with Label Quality Disparity in
  Federated Learning}.
\newblock
\newblock
\showeprint[arxiv]{2001.11359}~[cs.LG]


\bibitem[DeVries and Taylor(2017)]%
        {cutout}
\bibfield{author}{\bibinfo{person}{Terrance DeVries} {and}
  \bibinfo{person}{Graham~W. Taylor}.} \bibinfo{year}{2017}\natexlab{}.
\newblock \bibinfo{title}{Improved Regularization of Convolutional Neural
  Networks with Cutout}.
\newblock
\newblock
\urldef\tempurl%
\url{https://doi.org/10.48550/ARXIV.1708.04552}
\showDOI{\tempurl}


\bibitem[Duan et~al\mbox{.}(2022)]%
        {5}
\bibfield{author}{\bibinfo{person}{Shaoming Duan}, \bibinfo{person}{Chuanyi
  Liu}, \bibinfo{person}{Zhengsheng Cao}, \bibinfo{person}{Xiaopeng Jin}, {and}
  \bibinfo{person}{Peiyi Han}.} \bibinfo{year}{2022}\natexlab{}.
\newblock \showarticletitle{Fed-DR-Filter: Using global data representation to
  reduce the impact of noisy labels on the performance of federated learning}.
\newblock \bibinfo{journal}{\emph{Future Generation Computer Systems}}
  \bibinfo{volume}{137} (\bibinfo{year}{2022}), \bibinfo{pages}{336--348}.
\newblock
\showISSN{0167-739X}
\urldef\tempurl%
\url{https://doi.org/10.1016/j.future.2022.07.013}
\showDOI{\tempurl}


\bibitem[Fang and Ye(2022)]%
        {10}
\bibfield{author}{\bibinfo{person}{Xiuwen Fang} {and} \bibinfo{person}{Mang
  Ye}.} \bibinfo{year}{2022}\natexlab{}.
\newblock \showarticletitle{Robust Federated Learning With Noisy and
  Heterogeneous Clients}. In \bibinfo{booktitle}{\emph{Proceedings of the
  IEEE/CVF Conference on Computer Vision and Pattern Recognition}}.
  \bibinfo{pages}{10072--10081}.
\newblock


\bibitem[Goldberger and Ben-Reuven(2017)]%
        {bg1}
\bibfield{author}{\bibinfo{person}{Jacob Goldberger} {and}
  \bibinfo{person}{Ehud Ben-Reuven}.} \bibinfo{year}{2017}\natexlab{}.
\newblock \showarticletitle{Training deep neural-networks using a noise
  adaptation layer}. In \bibinfo{booktitle}{\emph{ICLR}}.
\newblock


\bibitem[Guo et~al\mbox{.}(2017)]%
        {calibration}
\bibfield{author}{\bibinfo{person}{Chuan Guo}, \bibinfo{person}{Geoff Pleiss},
  \bibinfo{person}{Yu Sun}, {and} \bibinfo{person}{Kilian~Q. Weinberger}.}
  \bibinfo{year}{2017}\natexlab{}.
\newblock \bibinfo{title}{On Calibration of Modern Neural Networks}.
\newblock
\newblock
\urldef\tempurl%
\url{https://doi.org/10.48550/ARXIV.1706.04599}
\showDOI{\tempurl}


\bibitem[He et~al\mbox{.}(2015)]%
        {resnet}
\bibfield{author}{\bibinfo{person}{Kaiming He}, \bibinfo{person}{Xiangyu
  Zhang}, \bibinfo{person}{Shaoqing Ren}, {and} \bibinfo{person}{Jian Sun}.}
  \bibinfo{year}{2015}\natexlab{}.
\newblock \bibinfo{title}{Deep Residual Learning for Image Recognition}.
\newblock
\newblock
\urldef\tempurl%
\url{https://arxiv.org/abs/1512.03385}
\showURL{%
\tempurl}


\bibitem[Helber et~al\mbox{.}(2018)]%
        {eurosat}
\bibfield{author}{\bibinfo{person}{Patrick Helber}, \bibinfo{person}{Benjamin
  Bischke}, \bibinfo{person}{Andreas Dengel}, {and} \bibinfo{person}{Damian
  Borth}.} \bibinfo{year}{2018}\natexlab{}.
\newblock \showarticletitle{Introducing Eurosat: A Novel Dataset and Deep
  Learning Benchmark for Land Use and Land Cover Classification}. In
  \bibinfo{booktitle}{\emph{IGARSS 2018 - 2018 IEEE International Geoscience
  and Remote Sensing Symposium}}. \bibinfo{pages}{204--207}.
\newblock
\urldef\tempurl%
\url{https://doi.org/10.1109/IGARSS.2018.8519248}
\showDOI{\tempurl}


\bibitem[Hendrycks et~al\mbox{.}(2018)]%
        {bengio}
\bibfield{author}{\bibinfo{person}{Dan Hendrycks}, \bibinfo{person}{Mantas
  Mazeika}, \bibinfo{person}{Duncan Wilson}, {and} \bibinfo{person}{Kevin
  Gimpel}.} \bibinfo{year}{2018}\natexlab{}.
\newblock \showarticletitle{Using Trusted Data to Train Deep Networks on Labels
  Corrupted by Severe Noise}. In \bibinfo{booktitle}{\emph{Advances in Neural
  Information Processing Systems}},
  \bibfield{editor}{\bibinfo{person}{S.~Bengio}, \bibinfo{person}{H.~Wallach},
  \bibinfo{person}{H.~Larochelle}, \bibinfo{person}{K.~Grauman},
  \bibinfo{person}{N.~Cesa-Bianchi}, {and} \bibinfo{person}{R.~Garnett}}
  (Eds.), Vol.~\bibinfo{volume}{31}. \bibinfo{publisher}{Curran Associates,
  Inc.}
\newblock
\urldef\tempurl%
\url{https://proceedings.neurips.cc/paper_files/paper/2018/file/ad554d8c3b06d6b97ee76a2448bd7913-Paper.pdf}
\showURL{%
\tempurl}


\bibitem[Hinton et~al\mbox{.}(2015)]%
        {hinton}
\bibfield{author}{\bibinfo{person}{Geoffrey Hinton}, \bibinfo{person}{Oriol
  Vinyals}, {and} \bibinfo{person}{Jeff Dean}.}
  \bibinfo{year}{2015}\natexlab{}.
\newblock \bibinfo{title}{Distilling the Knowledge in a Neural Network}.
\newblock
\newblock
\urldef\tempurl%
\url{https://doi.org/10.48550/ARXIV.1503.02531}
\showDOI{\tempurl}


\bibitem[Houle(2013)]%
        {lid}
\bibfield{author}{\bibinfo{person}{Michael~E. Houle}.}
  \bibinfo{year}{2013}\natexlab{}.
\newblock \showarticletitle{Dimensionality, Discriminability, Density and
  Distance Distributions}. In \bibinfo{booktitle}{\emph{2013 IEEE 13th
  International Conference on Data Mining Workshops}}.
  \bibinfo{pages}{468--473}.
\newblock
\urldef\tempurl%
\url{https://doi.org/10.1109/ICDMW.2013.139}
\showDOI{\tempurl}


\bibitem[Kairouz et~al\mbox{.}(2021)]%
        {OpenChallenges}
\bibfield{author}{\bibinfo{person}{Peter Kairouz}, \bibinfo{person}{H~Brendan
  McMahan}, \bibinfo{person}{Brendan Avent}, \bibinfo{person}{Aur{\'e}lien
  Bellet}, \bibinfo{person}{Mehdi Bennis}, \bibinfo{person}{Arjun~Nitin
  Bhagoji}, \bibinfo{person}{Kallista Bonawitz}, \bibinfo{person}{Zachary
  Charles}, \bibinfo{person}{Graham Cormode}, \bibinfo{person}{Rachel
  Cummings}, {et~al\mbox{.}}} \bibinfo{year}{2021}\natexlab{}.
\newblock \showarticletitle{Advances and open problems in federated learning}.
\newblock \bibinfo{journal}{\emph{Foundations and Trends{\textregistered} in
  Machine Learning}} \bibinfo{volume}{14}, \bibinfo{number}{1--2}
  (\bibinfo{year}{2021}), \bibinfo{pages}{1--210}.
\newblock


\bibitem[Konečný et~al\mbox{.}(2016)]%
        {fl}
\bibfield{author}{\bibinfo{person}{Jakub Konečný},
  \bibinfo{person}{H.~Brendan McMahan}, \bibinfo{person}{Felix~X. Yu},
  \bibinfo{person}{Peter Richtarik}, \bibinfo{person}{Ananda~Theertha Suresh},
  {and} \bibinfo{person}{Dave Bacon}.} \bibinfo{year}{2016}\natexlab{}.
\newblock \showarticletitle{Federated Learning: Strategies for Improving
  Communication Efficiency}. In \bibinfo{booktitle}{\emph{NIPS Workshop on
  Private Multi-Party Machine Learning}}.
\newblock
\urldef\tempurl%
\url{https://arxiv.org/abs/1610.05492}
\showURL{%
\tempurl}


\bibitem[Krizhevsky(2009)]%
        {cifar10}
\bibfield{author}{\bibinfo{person}{Alex Krizhevsky}.}
  \bibinfo{year}{2009}\natexlab{}.
\newblock \bibinfo{booktitle}{\emph{Learning multiple layers of features from
  tiny images}}.
\newblock \bibinfo{type}{{T}echnical {R}eport}.
\newblock


\bibitem[Leroy et~al\mbox{.}(2019)]%
        {KeywordSpottingIID}
\bibfield{author}{\bibinfo{person}{David Leroy}, \bibinfo{person}{Alice
  Coucke}, \bibinfo{person}{Thibaut Lavril}, \bibinfo{person}{Thibault
  Gisselbrecht}, {and} \bibinfo{person}{Joseph Dureau}.}
  \bibinfo{year}{2019}\natexlab{}.
\newblock \bibinfo{title}{Federated Learning for Keyword Spotting}.
\newblock
\newblock
\showeprint[arxiv]{1810.05512}~[eess.AS]


\bibitem[Li et~al\mbox{.}(2020b)]%
        {es2}
\bibfield{author}{\bibinfo{person}{Mingchen Li}, \bibinfo{person}{Mahdi
  Soltanolkotabi}, {and} \bibinfo{person}{Samet Oymak}.}
  \bibinfo{year}{2020}\natexlab{b}.
\newblock \showarticletitle{Gradient descent with early stopping is provably
  robust to label noise for overparameterized neural networks}. In
  \bibinfo{booktitle}{\emph{International conference on artificial intelligence
  and statistics}}. PMLR, \bibinfo{pages}{4313--4324}.
\newblock


\bibitem[Li et~al\mbox{.}(2020a)]%
        {9084352}
\bibfield{author}{\bibinfo{person}{Tian Li}, \bibinfo{person}{Anit~Kumar Sahu},
  \bibinfo{person}{Ameet Talwalkar}, {and} \bibinfo{person}{Virginia Smith}.}
  \bibinfo{year}{2020}\natexlab{a}.
\newblock \showarticletitle{Federated Learning: Challenges, Methods, and Future
  Directions}.
\newblock \bibinfo{journal}{\emph{IEEE Signal Processing Magazine}}
  \bibinfo{volume}{37}, \bibinfo{number}{3} (\bibinfo{year}{2020}),
  \bibinfo{pages}{50--60}.
\newblock
\urldef\tempurl%
\url{https://doi.org/10.1109/MSP.2020.2975749}
\showDOI{\tempurl}


\bibitem[Liu et~al\mbox{.}(2020)]%
        {energy_paper}
\bibfield{author}{\bibinfo{person}{Weitang Liu}, \bibinfo{person}{Xiaoyun
  Wang}, \bibinfo{person}{John~D. Owens}, {and} \bibinfo{person}{Yixuan Li}.}
  \bibinfo{year}{2020}\natexlab{}.
\newblock \bibinfo{title}{Energy-based Out-of-distribution Detection}.
\newblock
\newblock
\urldef\tempurl%
\url{https://doi.org/10.48550/ARXIV.2010.03759}
\showDOI{\tempurl}


\bibitem[Lukasik et~al\mbox{.}(2020)]%
        {ls_rate}
\bibfield{author}{\bibinfo{person}{Michal Lukasik}, \bibinfo{person}{Srinadh
  Bhojanapalli}, \bibinfo{person}{Aditya~Krishna Menon}, {and}
  \bibinfo{person}{Sanjiv Kumar}.} \bibinfo{year}{2020}\natexlab{}.
\newblock \bibinfo{title}{Does label smoothing mitigate label noise?}
\newblock
\newblock
\urldef\tempurl%
\url{https://doi.org/10.48550/ARXIV.2003.02819}
\showDOI{\tempurl}


\bibitem[McMahan et~al\mbox{.}(2017)]%
        {fedavg}
\bibfield{author}{\bibinfo{person}{Brendan McMahan}, \bibinfo{person}{Eider
  Moore}, \bibinfo{person}{Daniel Ramage}, \bibinfo{person}{Seth Hampson},
  {and} \bibinfo{person}{Blaise~Aguera y Arcas}.}
  \bibinfo{year}{2017}\natexlab{}.
\newblock \showarticletitle{Communication-efficient learning of deep networks
  from decentralized data}. In \bibinfo{booktitle}{\emph{Artificial
  intelligence and statistics}}. PMLR, \bibinfo{pages}{1273--1282}.
\newblock


\bibitem[Müller et~al\mbox{.}(2019)]%
        {ls}
\bibfield{author}{\bibinfo{person}{Rafael Müller}, \bibinfo{person}{Simon
  Kornblith}, {and} \bibinfo{person}{Geoffrey Hinton}.}
  \bibinfo{year}{2019}\natexlab{}.
\newblock \bibinfo{title}{When Does Label Smoothing Help?}
\newblock
\newblock
\urldef\tempurl%
\url{https://doi.org/10.48550/ARXIV.1906.02629}
\showDOI{\tempurl}


\bibitem[Northcutt et~al\mbox{.}(2019)]%
        {cl}
\bibfield{author}{\bibinfo{person}{Curtis~G. Northcutt}, \bibinfo{person}{Lu
  Jiang}, {and} \bibinfo{person}{Isaac~L. Chuang}.}
  \bibinfo{year}{2019}\natexlab{}.
\newblock \showarticletitle{Confident Learning: Estimating Uncertainty in
  Dataset Labels}.
\newblock  (\bibinfo{year}{2019}).
\newblock
\urldef\tempurl%
\url{https://doi.org/10.48550/ARXIV.1911.00068}
\showDOI{\tempurl}


\bibitem[Patrini et~al\mbox{.}(2016)]%
        {patrini}
\bibfield{author}{\bibinfo{person}{Giorgio Patrini},
  \bibinfo{person}{Alessandro Rozza}, \bibinfo{person}{Aditya Menon},
  \bibinfo{person}{Richard Nock}, {and} \bibinfo{person}{Lizhen Qu}.}
  \bibinfo{year}{2016}\natexlab{}.
\newblock \bibinfo{title}{Making Deep Neural Networks Robust to Label Noise: a
  Loss Correction Approach}.
\newblock
\newblock
\urldef\tempurl%
\url{https://doi.org/10.48550/ARXIV.1609.03683}
\showDOI{\tempurl}


\bibitem[Radford et~al\mbox{.}(2021)]%
        {radford2021learning}
\bibfield{author}{\bibinfo{person}{Alec Radford}, \bibinfo{person}{Jong~Wook
  Kim}, \bibinfo{person}{Chris Hallacy}, \bibinfo{person}{Aditya Ramesh},
  \bibinfo{person}{Gabriel Goh}, \bibinfo{person}{Sandhini Agarwal},
  \bibinfo{person}{Girish Sastry}, \bibinfo{person}{Amanda Askell},
  \bibinfo{person}{Pamela Mishkin}, \bibinfo{person}{Jack Clark},
  {et~al\mbox{.}}} \bibinfo{year}{2021}\natexlab{}.
\newblock \showarticletitle{Learning transferable visual models from natural
  language supervision}. In \bibinfo{booktitle}{\emph{International Conference
  on Machine Learning}}. PMLR, \bibinfo{pages}{8748--8763}.
\newblock


\bibitem[Ratner et~al\mbox{.}(2017)]%
        {weak_labels}
\bibfield{author}{\bibinfo{person}{Alexander Ratner},
  \bibinfo{person}{Stephen~H. Bach}, \bibinfo{person}{Henry Ehrenberg},
  \bibinfo{person}{Jason Fries}, \bibinfo{person}{Sen Wu}, {and}
  \bibinfo{person}{Christopher R{\'{e}}}.} \bibinfo{year}{2017}\natexlab{}.
\newblock \showarticletitle{Snorkel}.
\newblock \bibinfo{journal}{\emph{Proceedings of the {VLDB} Endowment}}
  \bibinfo{volume}{11}, \bibinfo{number}{3} (\bibinfo{date}{nov}
  \bibinfo{year}{2017}), \bibinfo{pages}{269--282}.
\newblock
\urldef\tempurl%
\url{https://doi.org/10.14778/3157794.3157797}
\showDOI{\tempurl}


\bibitem[Romero et~al\mbox{.}(2014)]%
        {fitnets}
\bibfield{author}{\bibinfo{person}{Adriana Romero}, \bibinfo{person}{Nicolas
  Ballas}, \bibinfo{person}{Samira~Ebrahimi Kahou}, \bibinfo{person}{Antoine
  Chassang}, \bibinfo{person}{Carlo Gatta}, {and} \bibinfo{person}{Yoshua
  Bengio}.} \bibinfo{year}{2014}\natexlab{}.
\newblock \bibinfo{title}{FitNets: Hints for Thin Deep Nets}.
\newblock
\newblock
\urldef\tempurl%
\url{https://doi.org/10.48550/ARXIV.1412.6550}
\showDOI{\tempurl}


\bibitem[Saeed et~al\mbox{.}(2021)]%
        {saeed2021contrastive}
\bibfield{author}{\bibinfo{person}{Aaqib Saeed}, \bibinfo{person}{David
  Grangier}, {and} \bibinfo{person}{Neil Zeghidour}.}
  \bibinfo{year}{2021}\natexlab{}.
\newblock \showarticletitle{Contrastive learning of general-purpose audio
  representations}. In \bibinfo{booktitle}{\emph{ICASSP 2021-2021 IEEE
  International Conference on Acoustics, Speech and Signal Processing
  (ICASSP)}}. IEEE, \bibinfo{pages}{3875--3879}.
\newblock


\bibitem[Shor and Venugopalan(2022)]%
        {shor2022trillsson}
\bibfield{author}{\bibinfo{person}{Joel Shor} {and} \bibinfo{person}{Subhashini
  Venugopalan}.} \bibinfo{year}{2022}\natexlab{}.
\newblock \showarticletitle{TRILLsson: Distilled Universal Paralinguistic
  Speech Representations}.
\newblock \bibinfo{journal}{\emph{arXiv preprint arXiv:2203.00236}}
  (\bibinfo{year}{2022}).
\newblock


\bibitem[Song et~al\mbox{.}(2020)]%
        {noises1}
\bibfield{author}{\bibinfo{person}{Hwanjun Song}, \bibinfo{person}{Minseok
  Kim}, \bibinfo{person}{Dongmin Park}, \bibinfo{person}{Yooju Shin}, {and}
  \bibinfo{person}{Jae-Gil Lee}.} \bibinfo{year}{2020}\natexlab{}.
\newblock \bibinfo{title}{Learning from Noisy Labels with Deep Neural Networks:
  A Survey}.
\newblock
\newblock
\urldef\tempurl%
\url{https://doi.org/10.48550/ARXIV.2007.08199}
\showDOI{\tempurl}


\bibitem[Tagliasacchi et~al\mbox{.}(2019)]%
        {Model}
\bibfield{author}{\bibinfo{person}{Marco Tagliasacchi}, \bibinfo{person}{Beat
  Gfeller}, \bibinfo{person}{F{\'e}lix de~Chaumont Quitry}, {and}
  \bibinfo{person}{Dominik Roblek}.} \bibinfo{year}{2019}\natexlab{}.
\newblock \showarticletitle{Self-supervised audio representation learning for
  mobile devices}.
\newblock \bibinfo{journal}{\emph{arXiv preprint arXiv:1905.11796}}
  (\bibinfo{year}{2019}).
\newblock


\bibitem[Tan et~al\mbox{.}(2021)]%
        {tan2021co}
\bibfield{author}{\bibinfo{person}{Cheng Tan}, \bibinfo{person}{Jun Xia},
  \bibinfo{person}{Lirong Wu}, {and} \bibinfo{person}{Stan~Z Li}.}
  \bibinfo{year}{2021}\natexlab{}.
\newblock \showarticletitle{Co-learning: Learning from noisy labels with
  self-supervision}. In \bibinfo{booktitle}{\emph{Proceedings of the 29th ACM
  International Conference on Multimedia}}. \bibinfo{pages}{1405--1413}.
\newblock


\bibitem[Trockman and Kolter(2022)]%
        {convmixer}
\bibfield{author}{\bibinfo{person}{Asher Trockman} {and}
  \bibinfo{person}{J.~Zico Kolter}.} \bibinfo{year}{2022}\natexlab{}.
\newblock \bibinfo{title}{Patches Are All You Need?}
\newblock
\newblock
\urldef\tempurl%
\url{https://doi.org/10.48550/ARXIV.2201.09792}
\showDOI{\tempurl}


\bibitem[Tsouvalas et~al\mbox{.}(2022)]%
        {fedstar}
\bibfield{author}{\bibinfo{person}{Vasileios Tsouvalas}, \bibinfo{person}{Aaqib
  Saeed}, {and} \bibinfo{person}{Tanir Ozcelebi}.}
  \bibinfo{year}{2022}\natexlab{}.
\newblock \showarticletitle{Federated Self-Training for Semi-Supervised Audio
  Recognition}.
\newblock \bibinfo{journal}{\emph{ACM Trans. Embed. Comput. Syst.}}
  (\bibinfo{date}{feb} \bibinfo{year}{2022}).
\newblock
\showISSN{1539-9087}
\urldef\tempurl%
\url{https://doi.org/10.1145/3520128}
\showDOI{\tempurl}


\bibitem[Wang et~al\mbox{.}(2019)]%
        {scce}
\bibfield{author}{\bibinfo{person}{Yisen Wang}, \bibinfo{person}{Xingjun Ma},
  \bibinfo{person}{Zaiyi Chen}, \bibinfo{person}{Yuan Luo},
  \bibinfo{person}{Jinfeng Yi}, {and} \bibinfo{person}{James Bailey}.}
  \bibinfo{year}{2019}\natexlab{}.
\newblock \showarticletitle{Symmetric cross entropy for robust learning with
  noisy labels}. In \bibinfo{booktitle}{\emph{IEEE International Conference on
  Computer Vision}}.
\newblock


\bibitem[{Warden}(2018)]%
        {spcm}
\bibfield{author}{\bibinfo{person}{P. {Warden}}.}
  \bibinfo{year}{2018}\natexlab{}.
\newblock \showarticletitle{{Speech Commands: A Dataset for Limited-Vocabulary
  Speech Recognition}}.
\newblock \bibinfo{journal}{\emph{ArXiv e-prints}} (\bibinfo{date}{April}
  \bibinfo{year}{2018}).
\newblock
\showeprint[arxiv]{1804.03209}~[cs.CL]


\bibitem[Wei et~al\mbox{.}(2021)]%
        {cifarn}
\bibfield{author}{\bibinfo{person}{Jiaheng Wei}, \bibinfo{person}{Zhaowei Zhu},
  \bibinfo{person}{Hao Cheng}, \bibinfo{person}{Tongliang Liu},
  \bibinfo{person}{Gang Niu}, {and} \bibinfo{person}{Yang Liu}.}
  \bibinfo{year}{2021}\natexlab{}.
\newblock \showarticletitle{Learning with Noisy Labels Revisited: {A} Study
  Using Real-World Human Annotations}.
\newblock \bibinfo{journal}{\emph{CoRR}}  \bibinfo{volume}{abs/2110.12088}
  (\bibinfo{year}{2021}).
\newblock
\showeprint[arXiv]{2110.12088}


\bibitem[Xia et~al\mbox{.}(2021)]%
        {es3}
\bibfield{author}{\bibinfo{person}{Xiaobo Xia}, \bibinfo{person}{Tongliang
  Liu}, \bibinfo{person}{Bo Han}, \bibinfo{person}{Chen Gong},
  \bibinfo{person}{Nannan Wang}, \bibinfo{person}{Zongyuan Ge}, {and}
  \bibinfo{person}{Yi Chang}.} \bibinfo{year}{2021}\natexlab{}.
\newblock \showarticletitle{Robust early-learning: Hindering the memorization
  of noisy labels}. In \bibinfo{booktitle}{\emph{International Conference on
  Learning Representations}}.
\newblock


\bibitem[Xiao et~al\mbox{.}(2017)]%
        {fmnist}
\bibfield{author}{\bibinfo{person}{Han Xiao}, \bibinfo{person}{Kashif Rasul},
  {and} \bibinfo{person}{Roland Vollgraf}.} \bibinfo{year}{2017}\natexlab{}.
\newblock \showarticletitle{Fashion-MNIST: a Novel Image Dataset for
  Benchmarking Machine Learning Algorithms}.
\newblock \bibinfo{journal}{\emph{CoRR}}  \bibinfo{volume}{abs/1708.07747}
  (\bibinfo{year}{2017}).
\newblock
\showeprint[arxiv]{1708.07747}


\bibitem[Xu et~al\mbox{.}(2022)]%
        {4}
\bibfield{author}{\bibinfo{person}{Jingyi Xu}, \bibinfo{person}{Zihan Chen},
  \bibinfo{person}{Tony Q.~S. Quek}, {and} \bibinfo{person}{Kai Fong~Ernest
  Chong}.} \bibinfo{year}{2022}\natexlab{}.
\newblock \bibinfo{title}{FedCorr: Multi-Stage Federated Learning for Label
  Noise Correction}.
\newblock
\newblock
\showeprint[arxiv]{2204.04677}~[cs.LG]


\bibitem[Yang et~al\mbox{.}(2021)]%
        {pathmnist}
\bibfield{author}{\bibinfo{person}{Jiancheng Yang}, \bibinfo{person}{Rui Shi},
  \bibinfo{person}{Donglai Wei}, \bibinfo{person}{Zequan Liu},
  \bibinfo{person}{Lin Zhao}, \bibinfo{person}{Bilian Ke},
  \bibinfo{person}{Hanspeter Pfister}, {and} \bibinfo{person}{Bingbing Ni}.}
  \bibinfo{year}{2021}\natexlab{}.
\newblock \showarticletitle{MedMNIST v2: A Large-Scale Lightweight Benchmark
  for 2D and 3D Biomedical Image Classification}.
\newblock  (\bibinfo{year}{2021}).
\newblock


\bibitem[Yang et~al\mbox{.}(2022b)]%
        {9}
\bibfield{author}{\bibinfo{person}{Miao Yang}, \bibinfo{person}{Hua Qian},
  \bibinfo{person}{Ximin Wang}, \bibinfo{person}{Yong Zhou}, {and}
  \bibinfo{person}{Hongbin Zhu}.} \bibinfo{year}{2022}\natexlab{b}.
\newblock \showarticletitle{Client Selection for Federated Learning With Label
  Noise}.
\newblock \bibinfo{journal}{\emph{IEEE Transactions on Vehicular Technology}}
  \bibinfo{volume}{71}, \bibinfo{number}{2} (\bibinfo{year}{2022}).
\newblock
\urldef\tempurl%
\url{https://doi.org/10.1109/TVT.2021.3131852}
\showDOI{\tempurl}


\bibitem[Yang et~al\mbox{.}(2022a)]%
        {2}
\bibfield{author}{\bibinfo{person}{Seunghan Yang}, \bibinfo{person}{Hyoungseob
  Park}, \bibinfo{person}{Junyoung Byun}, {and} \bibinfo{person}{Changick
  Kim}.} \bibinfo{year}{2022}\natexlab{a}.
\newblock \showarticletitle{Robust Federated Learning With Noisy Labels}.
\newblock \bibinfo{journal}{\emph{{IEEE} Intelligent Systems}}
  \bibinfo{volume}{37}, \bibinfo{number}{2} (\bibinfo{date}{mar}
  \bibinfo{year}{2022}), \bibinfo{pages}{35--43}.
\newblock
\urldef\tempurl%
\url{https://doi.org/10.1109/mis.2022.3151466}
\showDOI{\tempurl}


\bibitem[Yang et~al\mbox{.}(2018)]%
        {KeywordSuggestion}
\bibfield{author}{\bibinfo{person}{Timothy Yang}, \bibinfo{person}{Galen
  Andrew}, \bibinfo{person}{Hubert Eichner}, \bibinfo{person}{Haicheng Sun},
  \bibinfo{person}{Wei Li}, \bibinfo{person}{Nicholas Kong},
  \bibinfo{person}{Daniel Ramage}, {and} \bibinfo{person}{Françoise
  Beaufays}.} \bibinfo{year}{2018}\natexlab{}.
\newblock \bibinfo{title}{Applied Federated Learning: Improving Google Keyboard
  Query Suggestions}.
\newblock
\newblock
\showeprint[arxiv]{1812.02903}~[cs.LG]


\bibitem[Zeng et~al\mbox{.}(2022)]%
        {7}
\bibfield{author}{\bibinfo{person}{Bixiao Zeng}, \bibinfo{person}{Xiaodong
  Yang}, \bibinfo{person}{Yiqiang Chen}, \bibinfo{person}{Hanchao Yu}, {and}
  \bibinfo{person}{Yingwei Zhang}.} \bibinfo{year}{2022}\natexlab{}.
\newblock \showarticletitle{CLC: A Consensus-Based Label Correction Approach in
  Federated Learning}.
\newblock  \bibinfo{volume}{13}, \bibinfo{number}{5}, Article
  \bibinfo{articleno}{75} (\bibinfo{date}{jun} \bibinfo{year}{2022}),
  \bibinfo{numpages}{23}~pages.
\newblock
\showISSN{2157-6904}
\urldef\tempurl%
\url{https://doi.org/10.1145/3519311}
\showDOI{\tempurl}


\bibitem[Zhang et~al\mbox{.}(2018)]%
        {mixup}
\bibfield{author}{\bibinfo{person}{Hongyi Zhang}, \bibinfo{person}{Moustapha
  Cisse}, \bibinfo{person}{Yann~N. Dauphin}, {and} \bibinfo{person}{David
  Lopez-Paz}.} \bibinfo{year}{2018}\natexlab{}.
\newblock \showarticletitle{mixup: Beyond Empirical Risk Minimization}. In
  \bibinfo{booktitle}{\emph{International Conference on Learning
  Representations}}.
\newblock
\urldef\tempurl%
\url{https://openreview.net/forum?id=r1Ddp1-Rb}
\showURL{%
\tempurl}


\bibitem[Zhang et~al\mbox{.}(2023)]%
        {6}
\bibfield{author}{\bibinfo{person}{Jinghui Zhang}, \bibinfo{person}{Dingyang
  Lv}, \bibinfo{person}{Qiangsheng Dai}, \bibinfo{person}{Fa Xin}, {and}
  \bibinfo{person}{Fang Dong}.} \bibinfo{year}{2023}\natexlab{}.
\newblock \showarticletitle{Noise-Aware Local Model Training Mechanism for
  Federated Learning}.
\newblock \bibinfo{journal}{\emph{ACM Trans. Intell. Syst. Technol.}}
  (\bibinfo{date}{may} \bibinfo{year}{2023}).
\newblock
\showISSN{2157-6904}
\urldef\tempurl%
\url{https://doi.org/10.1145/3591363}
\showDOI{\tempurl}


\bibitem[Zhu et~al\mbox{.}(2021)]%
        {knn_paper}
\bibfield{author}{\bibinfo{person}{Zhaowei Zhu}, \bibinfo{person}{Zihao Dong},
  {and} \bibinfo{person}{Yang Liu}.} \bibinfo{year}{2021}\natexlab{}.
\newblock \bibinfo{title}{Detecting Corrupted Labels Without Training a Model
  to Predict}.
\newblock
\newblock
\urldef\tempurl%
\url{https://doi.org/10.48550/ARXIV.2110.06283}
\showDOI{\tempurl}


\end{thebibliography}
}

\newpage
\appendix
\section*{APPENDIX}
\begin{table*}[!htbp]
    \tiny \centering 
    \caption{\tiny{Performance evaluation for~\method's AKD approach, while globally aggregated model’s outputs are exploited as source of supervision. Average accuracy over three distinct trials on test set is reported. Federated parameters are set to $R$=200, $M$=30, $F$=80\%, $E$=1, $q$=80\%, and $\sigma$=25\%. Emb. denotes embedding.}\label{tab:main_res_extra}}
    \resizebox{0.85\hsize}{!}{%
        \begin{tabular}{lllccccccccc}
            \toprule
            \multicolumn{3}{l}{\textbf{Noise ($n_l$)}}                                       & 0.0 & \multicolumn{4}{c}{0.4} & \multicolumn{4}{c}{0.7} \\ \cmidrule[0.3pt](lr{.75em}){5-8} \cmidrule[0.3pt](lr{.75em}){9-12}
            \multicolumn{3}{l}{\textbf{Sparsity ($n_s$)}}                                    & 0.0 & 0.0 & 0.4 & 0.7 & 1.0 & 0.0 & 0.4 & 0.7 & 1.0 \\ \midrule
            \multirow{2}{*}{\textbf{CIFAR-10}}   
                                        & FedAvg &                                   & 78.52 & 68.77 & 67.05 & 67.31 & 67.92 & 57.65 & 56.94 & 56.81 & 63.54 \\
                                        & \multirow{2}{*}{\method} & AKD (Emb.)     & \textbf{83.55} & \textbf{69.72} & \textbf{69.18} & \textbf{70.26} & \textbf{70.55} & \textbf{68.43} & \textbf{68.06} & \textbf{69.74} & \textbf{68.39} \\
                                        &   & AKD (Logits)   & 78.54 & 63.42 & 62.79 & 67.45 & 68.63 & 54.09 & 53.38 & 59.14 & 64.73 \\ 
                                        \midrule
            \multirow{2}{*}{\textbf{\begin{tabular}[c]{l}Fashion\\MNIST\end{tabular}}}    
                                        & FedAvg &                                      & 86.43 & 82.05 & 83.24 & 83.35 & 81.07 & 58.55 & 56.06 & 57.11 & 59.37 \\
                                        &  \multirow{2}{*}{\method} & AKD (Emb.)       & \textbf{86.91} & \textbf{83.97} & \textbf{84.06} & \textbf{83.53} & \textbf{82.35} & \textbf{80.77} & \textbf{78.63} & \textbf{80.41} & \textbf{78.64} \\
                                        &   & AKD (Logits)     & 85.82 & 79.11 & 80.82 & 80.28 & 80.82 & 73.39 & 68.57 & 71.94 & 78.61 \\                                       
                                        \midrule
            \multirow{2}{*}{\textbf{\begin{tabular}[c]{l}Path\\MNIST\end{tabular}}}  
                                        & FedAvg &                                   & 87.05 & 78.82 & 77.06 & 76.68 & 77.03 & 54.74 & 52.49 & 53.22 & 58.61 \\
                                        & \multirow{2}{*}{\method} & AKD (Emb.)     & \textbf{87.82} & \textbf{86.42} & \textbf{82.83} & \textbf{81.46} & \textbf{83.26} & \textbf{79.94} & \textbf{78.63} & \textbf{78.31} & \textbf{76.02} \\
                                        & & AKD (Logits)   & 85.98 & 77.77 & 79.94 & 75.38 & 80.87 & 57.56 & 59.36 & 49.36 & 60.04 \\
                                        \midrule
            \multirow{2}{*}{\textbf{\begin{tabular}[c]{l}Speech\\Commands\end{tabular}}}   
                                        & FedAvg &                                   & 96.31 & 81.83 & 82.53 & 82.44 & 80.33 & 72.34 & 70.34 & 70.89 & 72.39 \\
                                        & \multirow{2}{*}{\method} & AKD (Emb.)     & 84.82 & \textbf{86.42} & \textbf{84.83} & \textbf{85.46} & \textbf{83.26} & \textbf{79.94} & \textbf{76.63} & \textbf{78.31} & \textbf{76.02} \\
                                        & & AKD (Logits)   & 95.69 & 80.39 & 81.07 & 82.56 & 81.64 & 73.11 & 72.74 & 74.14 & 69.67 \\

        \bottomrule
    \end{tabular}%
    }
    \vspace{-6pt}
\end{table*}

\end{document}